\documentclass[10pt,journal,compsoc]{IEEEtran}
\ifCLASSOPTIONcompsoc
  \usepackage[nocompress]{cite}
\else
  \usepackage{cite}
\fi
\usepackage{hyperref}
\usepackage{epsfig}
\usepackage{graphicx}
\usepackage{amsmath}
\usepackage{amssymb}
\usepackage{stfloats}
\usepackage{booktabs}
\usepackage{array}
\usepackage{algorithm}
\usepackage{algorithmic}
\usepackage{amsmath} 
\usepackage{amssymb}
\usepackage{bbding}
\usepackage{wasysym}
\usepackage{pifont}
\usepackage{graphicx}
\usepackage{subfigure}
\newcommand{\xmark}{\ding{55}}%

\ifCLASSINFOpdf
\else
\fi
\begin{document}
%
\title{Boosting Facial Expression Recognition by A Semi-Supervised Progressive Teacher}
%
%
%
%

\author{Jing~Jiang and
        Weihong~Deng,~\IEEEmembership{Member,~IEEE}
\IEEEcompsocitemizethanks{\IEEEcompsocthanksitem The authors are with the Pattern Recognition and Intelligent System Laboratory, School of Artificial Intelligence, Beijing University of Posts and Telecommunications, Beijing, 100876, China. Weihong Deng is the corresponding author.
E-mail:\{jiangjing1998,whdeng\} @bupt.edu.cn.}
\IEEEcompsocitemizethanks{\IEEEcompsocthanksitem This work was supported by the National Natural Science Foundation of China under Grants No. 61871052}
}

%
%

\markboth{Journal of \LaTeX\ Class Files,~Vol.~14, No.~8, August~2015}%
{Shell \MakeLowercase{\textit{et al.}}: Bare Demo of IEEEtran.cls for Computer Society Journals}
%



\IEEEtitleabstractindextext{%
\begin{abstract}
  In this paper, we aim to improve the performance of in-the-wild Facial Expression Recognition (FER) by exploiting semi-supervised learning. Large-scale labeled data and deep learning methods have greatly improved the performance of image recognition. However, the performance of FER is still not ideal due to the lack of training data and incorrect annotations (e.g., label noises). Among existing in-the-wild FER datasets, reliable ones contain insufficient data to train robust deep models while large-scale ones are annotated in lower quality. To address this problem, we propose a semi-supervised learning algorithm named Progressive Teacher (PT) to utilize reliable FER datasets as well as large-scale unlabeled expression images for effective training. On the one hand, PT introduces semi-supervised learning method to relieve the shortage of data in FER. On the other hand, it selects useful labeled training samples automatically and progressively to alleviate label noise. PT uses selected clean labeled data for computing the supervised classification loss and unlabeled data for unsupervised consistency loss. Experiments on widely-used databases RAF-DB and FERPlus validate the effectiveness of our method, which achieves state-of-the-art performance with accuracy of 89.57\% on RAF-DB. Additionally, when the synthetic noise rate reaches even 30\%, the performance of our PT algorithm only degrades by 4.37\%.
\end{abstract}

}

\maketitle

\IEEEdisplaynontitleabstractindextext

%
\IEEEpeerreviewmaketitle

\IEEEraisesectionheading{\section{Introduction}\label{sec:introduction}}

%
%
%
%
Facial expression is one of the most natural and universal way for human beings to convey their emotional states and intensions. Facial expression recognition (FER) has become a research hotspot in the field of computer vision since it's significant for human-computer interaction applications, such as remote education and driving fatigue monitoring devices. Researches on FER so far have mostly focused on discrete basic expression categories, i.e., anger, disgust, fearful, happy, sad and surprised. Recently, compound expressions have also been considered to describe more fine-grained emotions. According to data sources, databases for FER can be divided into two groups, which are lab-controlled and in-the-wild ones respectively. Lab-controlled expression datasets are conducted under the laboratory environment that subjects are asked to make specific expressions. This leads to accurate annotations but insufficient data. In-the-wild datasets consist of images collected from the Internet and manual annotations, so they usually have larger data quantity than lab-controlled ones due to easier data collection. However, it's time-consuming and resource-consuming to annotate a large-scale reliable dataset, which causes the contradiction between data quality and quantity. 

Early studies use lab-controlled datasets and hand-crafted features such as local binary patterns (LBP), histogram of gradient (HoG), and scale invariant feature transform (SIFT) to extract discriminative information and then train classifiers. This seems to work well because of the simple patterns of data. But the classifier will have poor performance on unseen data, especially images in real scenarios. Afterwards, significant progress has been made towards improving the performance of FER, especially researches based on deep learning method\cite{ding2017facenet2expnet,zeng2018facial,li2018reliable,wang2020suppressing}. With the development of deep learning, Convolutional Neural Networks (CNNs) remarkably benefit image recognition task. Large amount of high-quality data is crucial for training a robust CNN model. Considering training data, in-the-wild datasets\cite{li2017reliable,mollahosseini2017affectnet,barsoum2016training} are more suitable for FER in real-world condition than lab-controlled ones\cite{lucey2010extended,taini2008facial,valstar2010induced}. On the one hand, in-the-wild datasets usually have more training samples. On the other hand, more complex expression diversity makes the model more robust and adaptive. Even so, real-world FER still faces some challenges. Firstly, in-the-wild datasets may still have insufficient samples to train a robust deep neural network. Secondly, it's easy to collect expression-related images from Internet but it's hard to annotate them accurately. So the datasets may have inconsistent labels or incorrect labels (noisy labels) due to the uncertainty of samples\cite{wang2020suppressing}. These two issues cause over-fitting problem and thus limit the recognition performance, especially for data-driven deep learning based FER.

\begin{figure}[h]
\centering
\includegraphics[width=3.5in]{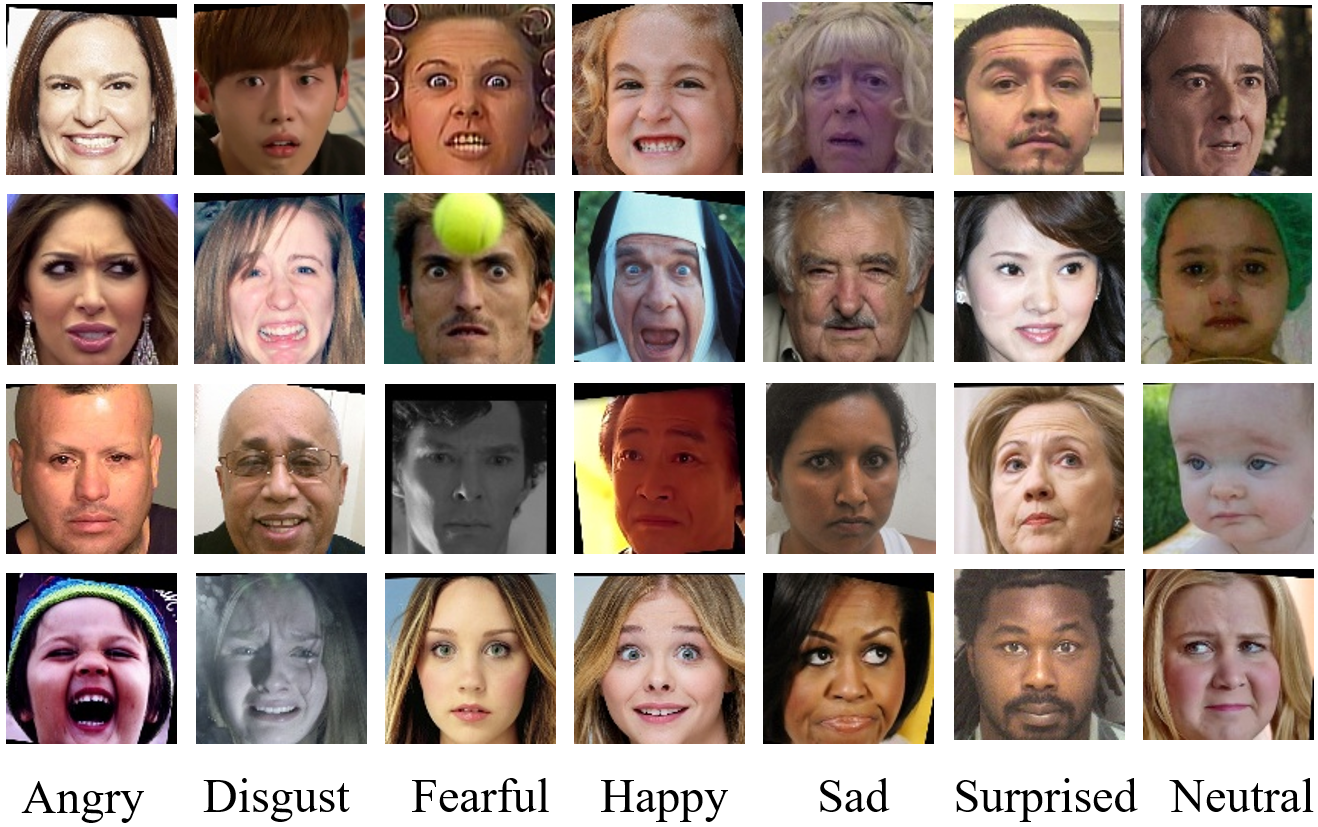}
\caption{Examples of not accurately annotated images in AffectNet.}
\label{AffectNet_wrong}
\end{figure}

Considering existing expression databases, we aim to fully utilize the reliable in-the-wild dataset as well as exploiting the effect of large-scale dataset with poor annotations. For example, AffectNet\cite{mollahosseini2017affectnet} is by far the largest database for FER which contains more than one millions of expression-related images. Among them about 280,000 training images are annotated into seven basic expressions (neutral is also considered), but each image is labeled by only one human-encoder so that the quality of annotation is not ensured. Therefore, the deep model trained with this dataset may have poor generalization performance. We show some apparently mislabeled samples in AffectNet in Fig.\ref{AffectNet_wrong}. Reliable in-the-wild FER datasets such as RAF-DB\cite{li2017reliable} and FERPlus\cite{barsoum2016training} contains much less data which also limits the strength of deep CNN. RAF-DB contains 12,271 training samples and each of them is labeled by 40 independent individuals. FERPlus consists of more than 20,000 training data and each of them is annotated by 10 experts. Label noises are relatively rare in these two datasets. We list the data distribution of above three datasets in Table\ref{number_3}.

\begin{table}[htbp]
    \centering
    \caption{\label{number_3}The data distribution of RAF-DB, FERPlus and AffectNet among seven expressions.}
    \setlength{\tabcolsep}{1.1mm}
    \small{
    \begin{tabular}{cccccccc}
    \toprule
    &Ang&Dis&Fea&Hap&Neu&Sad&Sur\\
    \midrule
    RAF-DB&705&717&281&4772&2524&1982&1290\\
    FERPlus&2399&175&648&7410&9365&3403&3378\\
    AffectNet&24882&3803&6378&134415&74874&25459&14090\\
    \bottomrule
    \end{tabular}}
\end{table}

Since it's difficult and time-consuming to annotate a large-scale in-the-wild FER dataset accurately, we can utilize the large amount of face images without annotations to help train a more robust model. Semi-supervised learning points out that the classifier which uses auxiliary unlabeled data can outperform that only uses small-size labeled ones. Therefore, we intend to take maximum advantage of the large-scale expression images in AffectNet as unlabeled data to boost the recognition performance. 

\begin{figure}[h]
    \centering
    \includegraphics[width=3.5in]{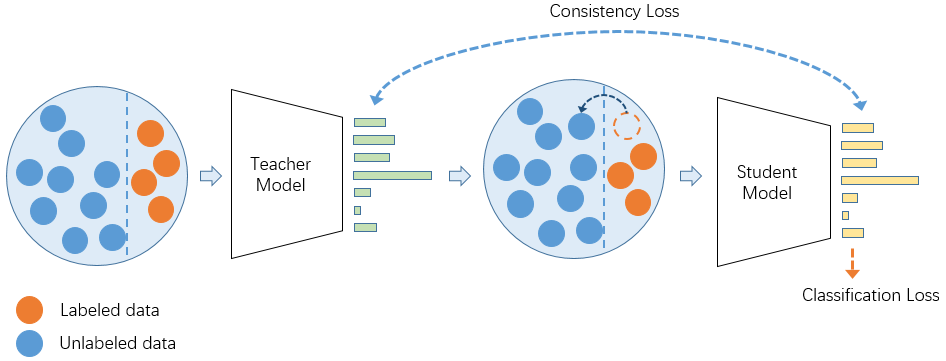}
    \caption{An illustration of Progressive Teacher.}
    \label{semi}
\end{figure}

To this end, we propose a semi-supervised deep learning framework named Progressive Teacher (PT) to address the shortage of data and label noise simultaneously. Progressive Teacher adopts semi-supervised learning to relieve the lack of training data in FER that auxiliary large amount of unlabeled data are utilized to boost the performance. Inspired by semi-supervised algorithm Mean Teacher\cite{tarvainen2017mean}, we follow the architecture of teacher-student model and make the network work in semi-supervised manner. Different from previous works on FER \cite{li2020deeper,liu2020omni} that unlabeled data are used to pre-train the network or annotated automatically to enlarge training set, we use both labeled and unlabeled data for computing the overall loss in a unified framework. Fig.\ref{semi} shows how the large-scale unlabeled data are utilized. The teacher model has stronger learning ability and guides the training of student model. Student model improves by learning to produce consistent outputs with the teacher model besides optimizing the classification loss of labeled data. Considering it costs much to construct a reliable in-the-wild FER database, we expect the network to be robust to label noises so that it can work even in noisy datasets. However, when the labeled dataset contains certain noisy labels, the negative effect of overfitting will gradually accumulate in this teacher-student framework such as Mean Teacher. Our Progressive Teacher alleviate this phenomenon by selecting useful labeled training samples automatically and progressively to tackle label noises. As illustrated in Fig.\ref{semi}, in each iteration, the teacher model will filter out a portion of labeled samples with increasing rate which are believed to be noisy ones, so as to prevent the student model from learning from their inaccurate annotations. Those samples are thereby regarded as unlabeled and used for consistency loss. In this case, labeled data are effectively utilized to prevent from overfitting and unlabeled data also improve the generalization performance in the meanwhile.

Overall, our contributions can be summarized as follows,

1. We analyze existing in-the-wild FER databases and aim to tackle the problem that reliable datasets contain insufficient labeled samples to train a robust deep neural network. Therefore, a semi-supervised algorithm Progressive Teacher (PT) is proposed. PT utilizes auxiliary large-scale unlabeled expression images for effective training in a semi-supervised manner to boost the recognition accuracy. Different from Mean Teacher, the teacher not only provides better logits and guides student to produce similar outputs, but also selects potential clean samples for student models to learn. Additionally, We use two pairs of teacher-student models and adopt a cross-guidance mechanism.

2. To alleviate the performance drop caused by noisy labels in traditional semi-supervised methods such as Mean Teacher, the proposed PT prevents from overfitting noisy samples by selecting clean labeled samples automatically and progressively. 

3. We extensively validate our PT algorithm on in-the-wild datasets RAF-DB and FERPlus. The results indicate that our semi-supervised method improves the recognition performance. Remarkably, we achieve the highest accuracy of 89.57\% on RAF-DB as far as we know. We also evaluate the effect of PT to tackle label noises on synthetic noisy datasets. It alleviates the performance drop evidently. When noisy samples get more, the effect of PT gets more remarkable. The experiment on real-world noisy dataset AffectNet also indicates the effectiveness of PT.
\section{Related Work}
    In this section, we mainly discuss methods that are related to facial expression recognition, semi-supervised learning and how to tackle with label noise and face uncertainties.
    \subsection{Facial expression recognition}

    As pointed out in \cite{li2020deep}, deep learning based facial expression recognition can be performed by three major steps: data pre-processing, deep feature learning and deep feature classification. We get aligned face regions from original images in the pre-processing stage, and then feed the aligned images into a deep neural network, for example, Convolutional Neural Network(CNN), to obtain discriminative features and then classification will be conducted almost simultaneously. Deep learning based FER requires large amount of training samples to avoid the over-fitting problem. However, reliable in-the-wild FER databases, such as RAF-DB\cite{li2017reliable} and FERPlus\cite{barsoum2016training} usually have insufficient data to learn discriminative deep features. The annotation quality of the large-scale database AffectNet\cite{mollahosseini2017affectnet} is not ensured due to large data quantity, which limits the strength of CNN.

    To tackle the shortage of training data in FER, some works\cite{levi2015emotion,zhao2016peak} utilize face recognition datasets to pre-train the network and then fine-tune it on the expression dataset. To eliminate the effect of face-dominated information reserved in the pre-trained face network,  FaceNet2ExpNet\cite{ding2017facenet2expnet} was proposed to make the pre-trained model provide a good initialization. Considering utilizing more expression datasets to enlarge the amount of data, IPA2LT\cite{zeng2018facial} has found that the performance of FER can't be improved and even degrades by merging multiple datasets directly due to inconsistent annotation. To avoid this effect, IPA2LT uses multiple models trained on different datasets to discover the latent truth of input images. Recently, a omni-supervised FER baseline\cite{liu2020omni} was proposed to make use of auxiliary large-scale unlabeled images. It first trains a primitive model using a small number of labeled samples, then use it to select high confident unlabeled samples by feature-based similarity comparison. The enlarged dataset is proved to boost the recognition performance. It also adopts a dataset distillation strategy to distill and compress the useful knowledge from the selected auxiliary samples in order to reduce computation resources. \cite{happy2019weakly} proposed a weakly supervised learning technique, which uses unlabeled data with high confidence scores as labeled ones and train the network in a supervised way. Similar to \cite{liu2020omni}, we aim to exploit the effect of large-scale unlabeled images, but in an intrinsically semi-supervised manner. Our method is an extension of teacher-student model. \cite{georgescu2020teacher} also employ the teacher-student training strategy and it treats model trained on fully-visible faces as teacher and model trained on occluded faces as student to learn discriminative embeddings for classifying expressions under occlusion. Knowledge distillation and triplet loss are utilized for learning embeddings.
    
    \subsection{Semi-supervised Learning}

    The success of deep learning bases on large amount of well-annotated data. However, constructing a large-scale dataset with high-quality annotation is time-consuming and labor-expensive. Semi-supervised learning tackles the shortage of labeled training data and improves learning ability. As illustrated in \cite{van2020survey}, semi-supervised learning uses labelled as well as unlabeled data to perform certain learning tasks such as classification. Some methods\cite{lee2013pseudo,guo2019robust} which can be classes as self-training use base classifiers to obtain predictions for unlabeled data and these pseudo-labelled data are therefore used for supervised training. In \cite{lee2013pseudo}, unlabeled images are labeled as the most confident of their predictions. The unlabeled data are weighted to fit the training process since pseudo labels are not reliable enough in the beginning. \cite{guo2019robust} balances the influence of unlabeled data by giving pseudo label when the maximum prediction confidence is greater than a threshold. Co-training\cite{du2010does} and Tri-training\cite{zhou2005tri} are extensions of self-training which learn from multiple supervised classifiers. 
    Some methods \cite{rasmus2015semi,laine2016temporal,tarvainen2017mean} based on deep learning make the network work in semi-supervised fashion instead of giving the unlabeled data pseudo labels in advance. $\Pi$-model\cite{laine2016temporal} exploits unlabeled data by constraining the predictions of two models to be consistent under different data augmentation and dropout. The total loss consists of two parts, one is the classification loss of labeled data, the other is the consistency loss of all data without labels. Temporal ensembling\cite{laine2016temporal} is an extension of $\Pi$-model, it penalizes the difference in the network outputs at different points in time during the training process. It's an implicit teacher-student model which constructs a better target by ensembling the outputs during training to regularize the student model. Compared to \cite{laine2016temporal}, Mean teacher\cite{tarvainen2017mean} is an explicit teacher-student model which constructs a better teacher model by averaging the weights at each training iteration. 
    Inspired by \cite{tarvainen2017mean}, we also make a good teacher model by weight-averaging and regularize the student model by penalizing the difference in network outputs with teacher model. But we use two pairs of teacher-student models, assisted by a cross-guidance mechanism, to make it more robust.
    \subsection{Learning with Label Noise and Face Uncertainties}
    Noisy labels in training dataset may hamper the performance of trained CNNs and thus lots of works focus on learning with label noise. Some researches proposed robust loss functions \cite{ghosh2015making,charoenphakdee2019symmetric,ghosh2017robust,zhang2018generalized} so that even noisy labels are fed for training models, they will not decrease the performance. Other works aim to model the noise in order to  relabel, re-weight or remove them. Some methods\cite{vahdat2017toward,veit2017learning,tanaka2018joint,yi2019probabilistic,arazo2019unsupervised} identify and correct suspicious labels to their corresponding true class. Some methods\cite{liu2015classification,ren2018learning,shu2019meta,lee2018cleannet} try to assign importance weights to training samples. Usually clean samples will have larger weight value to reduce the influence of noisy labels. Some works\cite{reed2014training,chang2017active,malach2017decoupling,han2018co} aim to guide the network to choose clean instances for updating parameters. The co-teaching learning paradigm was proposed in \cite{han2018co} that two neural networks are trained simultaneously to teach each other with potential clean samples. In the field of face recognition, \cite{wu2018light} aims to remove potential noise and construct a clean dataset while \cite{zhong2019unequal} proposed a noise-robust loss function to tackle with incorrect identities. 

    Recently, the uncertainties in the FER task has been illustrated. Due to ambiguous facial expressions, low-quality facial images, inconsistent annotations and noisy labels, the network tends to fit these samples and thus causing over-fitting. Zeng et al.\cite{zeng2018facial} firstly take the inconsistent annotation problem among different datasets into FER and try to find the latent truth with multiple models. Wang et.al.\cite{wang2020suppressing} suppresses the uncertainties by sample reweighting and relabeling scheme. The network learns the uncertainty of samples with a self-attention mechanism and uncertain samples will have small importance weights and then be relabeled for training. Fan et.al.\cite{fan2020learning} estimate the uncertainty by comparing the distance between sample and its class center in the embedding space. Outliers will have small weight and then weighted softmax is utilized to suppress the impact of uncertain instances.

    Except utilizing large amount of unlabeled data to tackle the shortage of training data, our Progressive Teacher also takes the uncertainties of FER into account and makes the network robust to noisy labels. Studies on the memory effect of deep neural network show that deep networks will learn clean and easy instances in the early stage of training\cite{arpit2017closer}. Intuitively, the clean and easy samples have smaller loss values while noisy samples have larger ones. As training continues, they will gradually memorize and overfit those noisy labels. We adopt this training scheme in \cite{han2018co} to train a noise-robust model as well as using auxiliary unlabeled images.
\section{Progressive Teacher} 

\begin{figure*}[h]
        \centering
        \includegraphics[scale=0.5]{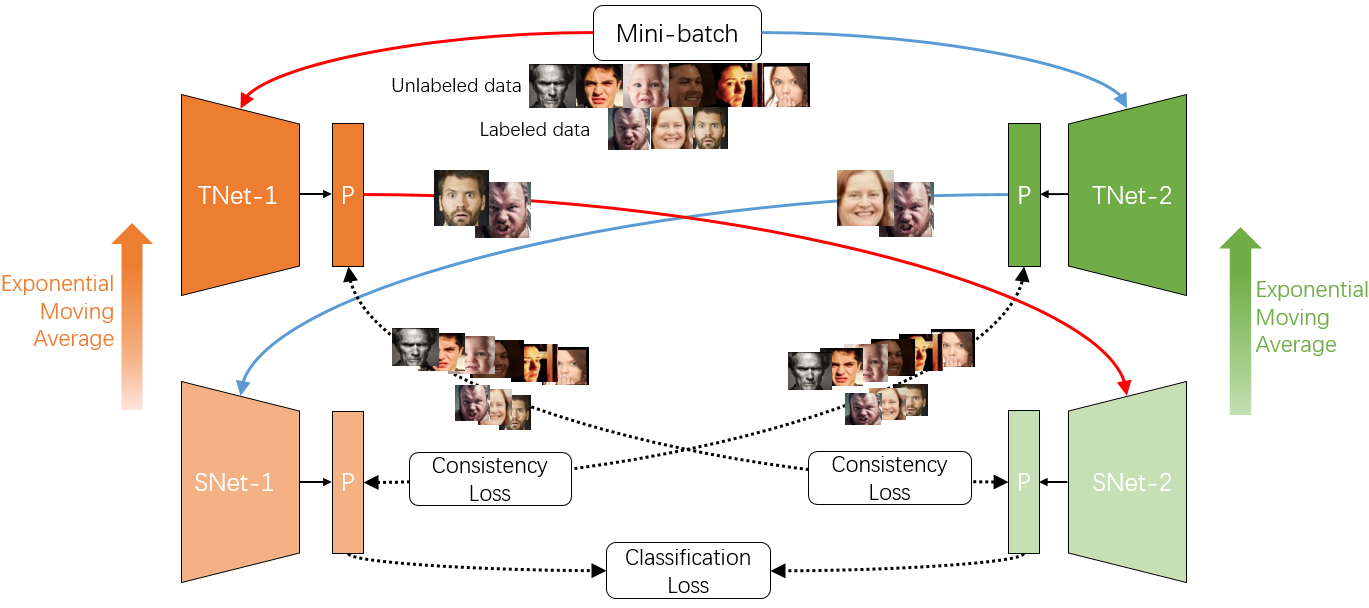}
        \caption{The pipeline of our semi-supervised algorithm Progressive Teacher. Both labeled and unlabeled face images are fed into two teacher models with better learning capability (TNets, TNet-1 and TNet-2) simultaneously. Then TNets will select a part of labeled samples, which are believed to be useful and clean, and then fed them into the student models (SNets, SNet-2 and SNet-1) in the other group. TNets share the same network architecture with SNets and derive from SNets by exponentially averaging their weights along training steps. The number of selected clean samples decreases progressively to meet the training process and avoid remembering potential noisy samples. In the meanwhile, large-scale unlabeled data works that when feeding them into networks, SNets improve themselves by learning to output consistent probability distributions with TNets. For each SNet, it will compute the classification loss of selected labeled samples and the consistency loss using all samples in an mini-batch.}
        \label{img1}
    \end{figure*}

    Our proposed Progressive Teacher is a typical teacher-student model, which means that the student model improves itself gradually with the guidance of teacher model. More specifically, the teacher model is the average of student model's weights during training process and has better performance, the student model is promoted by optimizing the classification loss of labeled data and, in the meanwhile, learning to produce consistent outputs with the teacher model. As a semi-supervised algorithm, unlabeled data as well as labeled ones are used to compute the later consistency loss by penalizing the difference in the outputs of two models with the same image. Under the guidance of teacher model, the student model improves increasingly and finally achieves similar learning ability. This semi-supervised training strategy is proved to work well tackling the shortage of data. However, when there are label noise in dataset, which is inevitable in FER task, the student model will gradually fit those noisy samples when optimizing the classification loss and then the performance will degrade. Additionally, this passive effect will accumulate in the teacher model, resulting in the decline of its guidance ability. Consequently, the overall performance will drop. Therefore, our Progressive Teacher utilizes a unified framework which is shown in Fig \ref{img1}, to tackle both the data shortage and noisy labels simultaneously. It abandons potential noisy samples automatically and progressively to make the student model learn pure useful knowledge. Compared to traditional teacher-student model like Mean Teacher\cite{tarvainen2017mean}, the teacher not only provide better logits and guides student to produce similar outputs, but also selects potential clean samples for student models to learn. Additionally, we use two pairs of teacher-student models and adopt a cross-guidance mechanism. The two groups complement each other and boost recognition.

    The two groups of teacher-student models share the same architecture but have different initialization. In each group, the student model (SNet) minimizes loss function and is optimized by stochastic gradient descent (SGD), the teacher model (TNet) derives from its student model by exponential average moving. Due to different initialization between two groups, they will feed each other with different samples in the early stage of training. Trained with different samples, the two student models differ and this variance will accumulate in their teachers. In this circle, we argue that the two groups complement each other and will learn better. Similar to Mean teacher, we regard the average model as teacher model which performs better and guides the student model. Differently, the teacher not only provides guidance, but also select clean samples for student to learn. When inputting samples into TNets, those with smaller loss (cross-entropy loss) values are believed to be clean and then conveyed to the SNets in the other group for further training. A cross-guidance mechanism is adopted to boost performance. For example, selected labeled clean samples from TNet-1 will be conveyed to SNet-2, then SNet-2 will compute the supervised classification loss of them and the unsupervised consistency loss with TNet1 to update network weights. Specifically, we use cross-entropy loss as classification loss and mean squared error (MSE) as consistency loss.

    Given labeled data set $L=\{(x_{i},y_{i})\}$ and unlabeled data set $U=\{x_j\}$, the teacher and student model use different data augmentations when inputting images. We denote the student model as $f(x,\theta,\eta)$ and teacher model as $f(x,\theta^{'},\eta^{'})$. 

    For convenience, the softmax of their outputs are also denoted as the same formulation.
    In each iteration with mini-batch  $B$, clean samples selected by TNet-1 and TNet-2 are denoted as $B_{1}$ and $B_{2}$ respectively. The ratio of selected clean samples is represented as $R(t)$ and changes its value along iteration.
    For SNet-1, the supervised cross-entropy loss $L_{s1}$ is denoted as following formulation:
    \begin{equation}
    L_{s1}=\frac{1}{\left| L\cap B_{2}\right|}\sum_{i\in L\cap B_{2}} \sum_{j=1}^{C} y_{i,j}\log f_{j}(x_{i},\theta_{1},\eta)
    \end{equation}
    in which $y_{i,j}$ is the $j$-th value of the ground truth label of the $i$-th sample and $f_{j}(\cdot)$ represents the $j$-th output value of softmax of student model. $C$ is the number of classes.

    The  unsupervised consistency loss is

    \begin{equation}
        L_{u1}=\frac{1}{\left|B\right|}\sum_{i\in B}\left\|f(x_{i},\theta_{2}^{'},\eta^{'})-f(x_{i},\theta_{1},\eta) \right \|^{2}
    \end{equation}

    Here \emph{unsupervised} means that only images are utilized, without their labels. So we use all samples for computing unsupervised loss, which means that labeled data are also included. 
    The overall loss for SNet-1 is

    \begin{equation}
        L_{1}=L_{s1}+\omega(t)L_{u1}
    \end{equation}
    
    in which $\omega(t)$ is unsupervised weight ramp-up function. Because the teacher model has poor guidance in the beginning of training, we should give the unsupervised loss a small weight first and increase it gradually. Similarly, the loss function for SNet-2 is denoted as
    \begin{equation}
        L_{2}=L_{s2}+\omega(t)L_{u2}
    \end{equation}

    in which
    \begin{equation}
        L_{s2}=\frac{1}{\left| L\cap B_{1}\right|}\sum_{i\in L\cap B_{1}} \sum_{j=1}^{C} y_{i,j}\log f_{j}(x_{i},\theta_{2},\eta)
    \end{equation}

    and
    \begin{equation}
        L_{u2}=\frac{1}{\left|B\right|}\sum_{i\in B}\left\|f(x_{i},\theta_{1}^{'},\eta^{'})-f(x_{i},\theta_{2},\eta) \right \|^{2}
    \end{equation}
    
    In each iteration, after the student models update their weights, the teacher models are updated by following exponential moving average way. $\alpha$ is a smoothing coefficient hyperparameter.

    \begin{equation}
        \theta_{t}^{'}=\alpha\theta_{t-1}^{'}+(1-\alpha)\theta_{t}
    \end{equation}

    As illustrated before, the network prefers to learn from clean samples and is not sensitive to noisy ones in the early stage of training, so we feed the SNets with more labeled samples in the early stage and then progressively reduce the amount to avoid picking noisy ones. Algorithm \ref{alg:Framwork} summarizes the learning process of our Progressive Teacher.
    \begin{algorithm}[htbp] 
      \caption{The pipeline of Progressive Teacher.}
      \label{alg:Framwork} 
      \begin{algorithmic}[1] 
      \REQUIRE ~~\\ 
      The labeled data sets L and unlabeled data sets U;\\
      TNets, $f(x,\theta_{i}^{'},\eta^{'})$, $i=1,2$;\\
      SNets, $f(x,\theta_{i},\eta)$, $i=1,2$
      \ENSURE ~~ 
      Parameters $\theta_{1}$, $\theta_{2}$, $\theta_{1}^{'}$, $\theta_{2}^{'}$\\
      Learning rate $\lambda$, cross-entropy loss l, $t=0$
      \FOR{epoch $i \in [1,N]$}
          \FOR{mini-batch $B$}
          \STATE $t=t+1$
          \STATE Select 100*R(t)\% small-loss samples by TNet-1:\\
           $B_{1}=\mathop{\arg\min}_{{B^{'}:\left|B^{'}\right|\ge R(t)\left|B\right|}}l(f(x,\theta_{1}^{'},\eta^{'}),B)$ \\
          \STATE Select 100*R(t)\% small-loss samples by TNet-2:\\
           $B_{2}=\mathop{\arg\min}\limits_{{B^{'}:\left|B^{'}\right|\ge R(t)\left|B\right|}}l(f(x,\theta_{2}^{'},\eta^{'}),B)$ 
          \STATE Update parameters in SNet-1:\\
          $\theta_{1,t}=\theta_{1,t-1}-\lambda\frac{\partial L_{1}}{\partial \theta_{1,t-1}}$
          \STATE Update parameters in SNet-2:\\
          $\theta_{2,t}=\theta_{2,t-1}-\lambda\frac{\partial L_{2}}{\partial \theta_{2,t-1}}$
          \STATE Update parameters in TNet-1:\\
          $\theta_{1,t}^{'}=\alpha\theta_{1,t-1}^{'}+(1-\alpha)\theta_{1,t}$
          \STATE Update parameters in TNet-2:\\
          $\theta_{2,t}^{'}=\alpha\theta_{2,t-1}^{'}+(1-\alpha)\theta_{2,t}$
          
          \ENDFOR
      \ENDFOR
      \end{algorithmic}
  \end{algorithm}
    We adopt two pairs of teacher-student models and use TNet from the other group to guide the SNet according to following considerations. Firstly, we use complementary TNets to select clean samples instead of SNets. Because SNets are optimized with SGD, the parameters update quickly and are more affected by noisy labels. TNets are more stable and learn better due to the accumulation along time. Therefore, small-loss samples selected by the later are more reliable. Secondly, we penalize the difference in outputs of SNet-1 (SNet-2) and TNet-2 (TNet-1). In semi-supervised learning, perturbation-based methods expect networks to produce consistent outputs when adding small noise to both inputs and models themselves. To this end, we can get larger perturbation in models when using TNet and SNet from different groups. Additionally, SNets can learn more information from the variance of two groups since they are initialized differently and trained with not exactly the same samples.
    As the student model finally achieves comparable performance with its teacher, we choose one of the two teacher models for testing after training is finished. 
    
\section{Experiments}
In this section, we evaluate our method on RAF-DB and FERPlus with using AffectNet as auxiliary unlabeled data. Firstly, we will show the efficiency of semi-supervised learning. To further demonstrate the effect of tackling noisy labels, we add different levels of label noises to the above two datasets since the original datasets are believed to be reliable approximately. Specifically, for symmetric noise, we randomly select certain ratio of training samples in each expression category and change their labels to others uniformly. Considering expression surprised shares high similarity with fearful since they often occur with opened mouth or widened eyes, which may confuse annotators in real scenario, we also add asymmetric noise that the two expressions are relabeled to each other. Additionally, we conduct experiment on real-world noisy dataset AffectNet.

\subsection{Datasets}
\textbf{RAF-DB}\cite{li2017reliable} is a large-scale in-the-wild facial expression database which contains about 30,000 great diverse facial images from thousands of individuals downloaded from the Internet. Images in RAF-DB were labeled by 315 human coders and the final annotations were determined through the crowdsourcing techniques. And each image was assured to be labeled by about 40 independent labelers. RAF-DB contains 12,271 training samples and 3,068 test samples annotated with seven basic emotional categories(neutral expression is also taken into account). Compound expressions labeled as 11 classes are also provided, which are not used in our experiment.

\textbf{FERPlus}\cite{barsoum2016training} is an extension of FER2013\cite{goodfellow2013challenges}. The large-scale and unconstrained database FER2013 was created and labeled automatically by the Google image search API. All images in FER2013 have been registered and resized to 48*48 pixels. 
FER2013 contains 28,709 training images, 3,589 validation images and 3,589 test images with seven expression labels. It's relabeled in 2016 by Microsoft with each image labeled by 10 individuals to consist 8 classes (contempt is added), thus has more reliable annotations. We only use seven basic emotional categories as RAF-DB.

\textbf{AffectNet}\cite{mollahosseini2017affectnet} contains more than 0.4 millions of labeled images which also includes contempt. The images are downloaded from Internet using three search engines and expression-related keywords. It's the largest dataset for FER currently. We choose the labeled images in 7-class basic expression categories, resulting in about 280,000 samples. In later experiment, we use these large amount of images as auxiliary unlabeled data without using their label information.

\subsection{Implementation}    
\begin{figure}[h]
\centering
\includegraphics[width=3.5in]{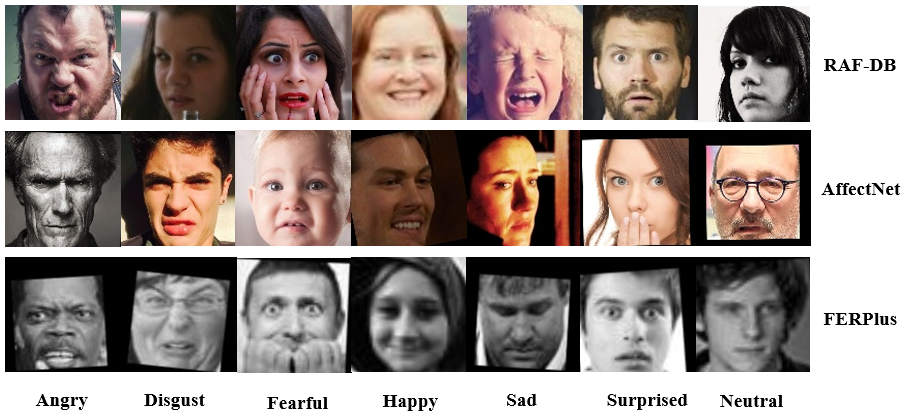}
\caption{Aligned images in three datasets.}
\label{image_example_2}
\end{figure}

For all images in RAF-DB, FERPlus and AffectNet, we detect face regions and locate their landmark points using MTCNN, then align them by similarity transformation. All images are further resized to 112*112 pixels, which are shown in Fig.\ref{image_example_2}. We use ResNet-18 as our backbone network, which is pre-trained on face recognition dataset CASIA-WebFace\cite{yi2014learning}. The dimension of feature is fixed to 512. All experiments are conducted under Pytorch framework, using one NVIDIA TITAN Xp GPU.

For Progressive Teacher, the student model will improve from fast to slow, so we should gradually increase the value of $\alpha$. In the early stage of training, $\alpha$ with small value makes the teacher model quickly forget previous inaccurate student weights. As training goes further, the teacher model benefits from long-term memory with larger $\alpha$. Specifically, $\alpha=min\{1-1/(1+iter),0.999\}$, in which iter stands for iterations. We train the model for $N$ epochs totally and N equals to 6 in experiment. The ratio of selected clean samples by teacher models 
\begin{equation}R(t)=1-r*min\{t/T,1\}
\end{equation},
in which r is set to 0.05 on original RAF-DB and FERPlus, $r^{'}+0.1$ respectively when adding symmetric noise ratio of $r^{'}$. When adding asymmetric noise to RAF-DB, r is set to 0.05 when noise ratio is 10\%, r equals to 10\% on other ratios. When adding asymmetric noise to FERPlus, r is set to 0.1 when noise ratio is 10\%, 20\%, 30\%, r equals to 0.15 on other ratios. t represents the current iteration and $T$ is the turning iteration and set to 300. $R(t)$ decrease progressively to fit the training progress of neural networks that they are able to filter out noises automatically in the beginning and gradually overfit. After $T$ iterations, $R(t)$ keeps the same. The minibatch size is fixed to 200, in which 1/4 are labeled data and others are unlabeled. This facilitates the converge process and makes it controllable. We use random horizontal flip as data augmentation and SGD with momentum as optimizer. The momentum is 0.9 and weight decay is 0.0005. The learning rate starts with 0.01 and decreases to 0.001 after 3 epochs. The unsupervised weight ramp-up function $\omega(t)=10*exp\{-5*(1-epoch/N)^2\}$.

\begin{figure*}[htbp]
    \centering
    \subfigure[Fine-tune]{\includegraphics[width=0.3\hsize]{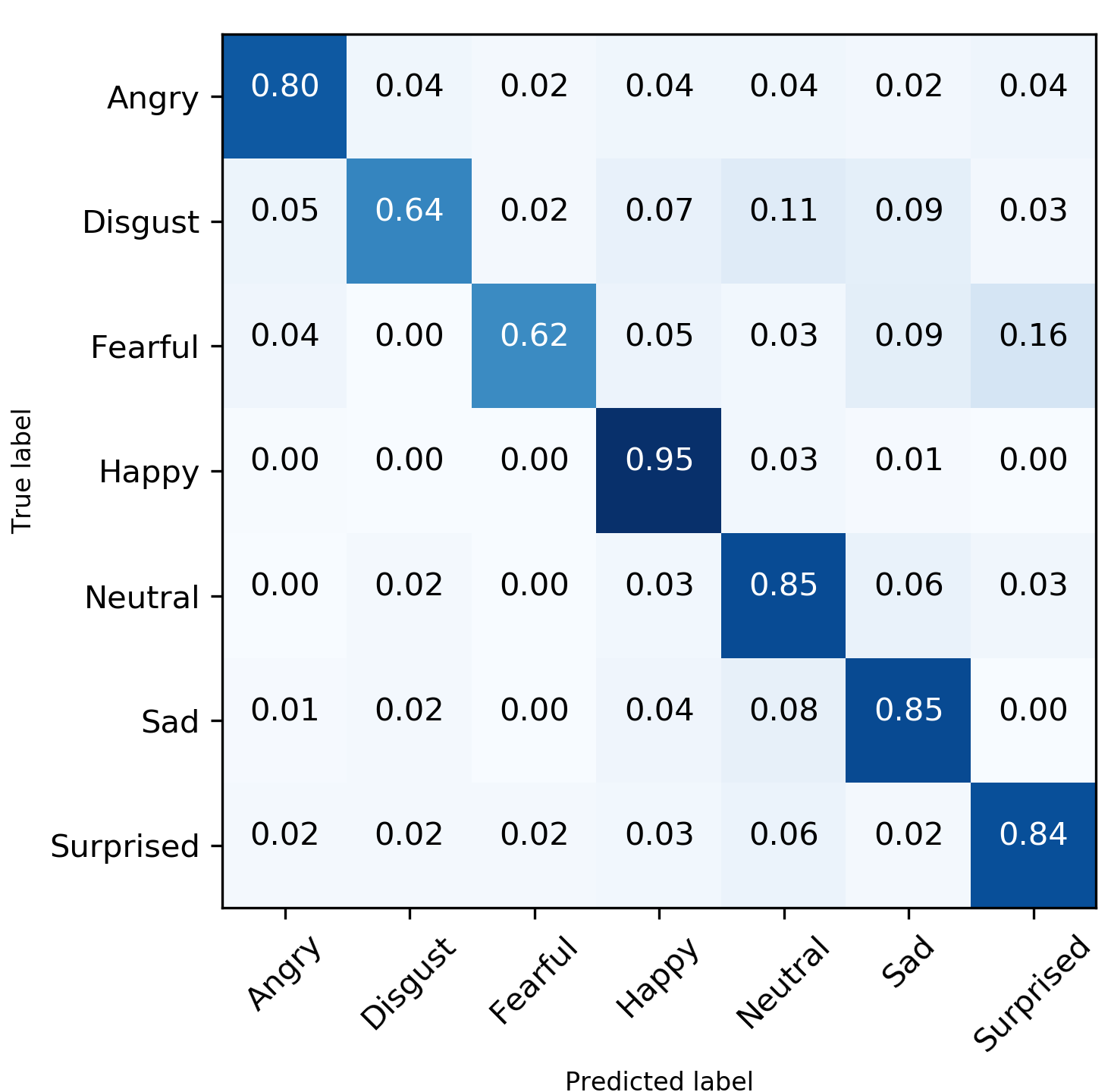}} 
    \subfigure[Mean Teacher]{\includegraphics[width=0.3\hsize]{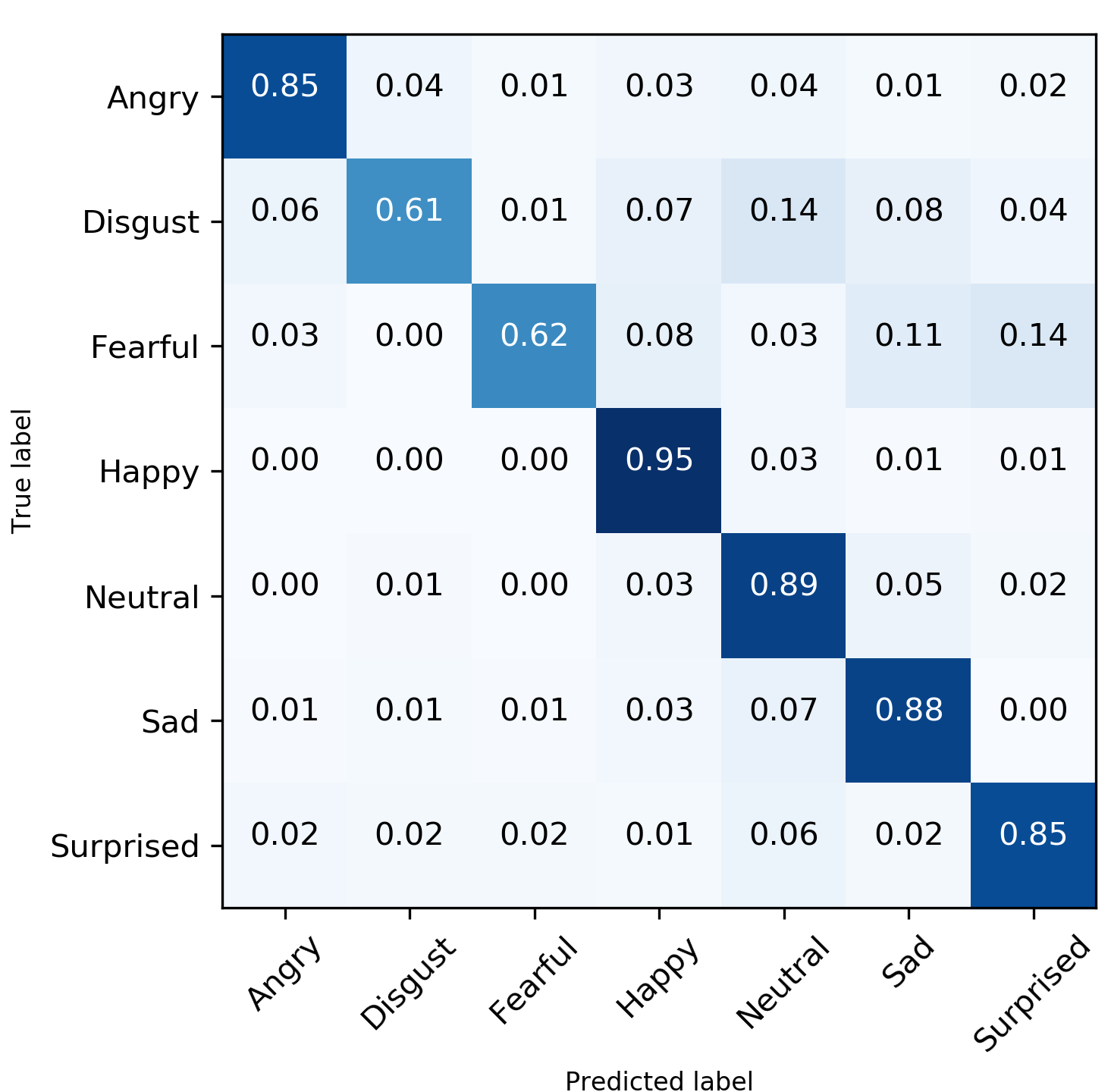}} 
    \subfigure[Progressive Teacher]{\includegraphics[width=0.3\hsize]{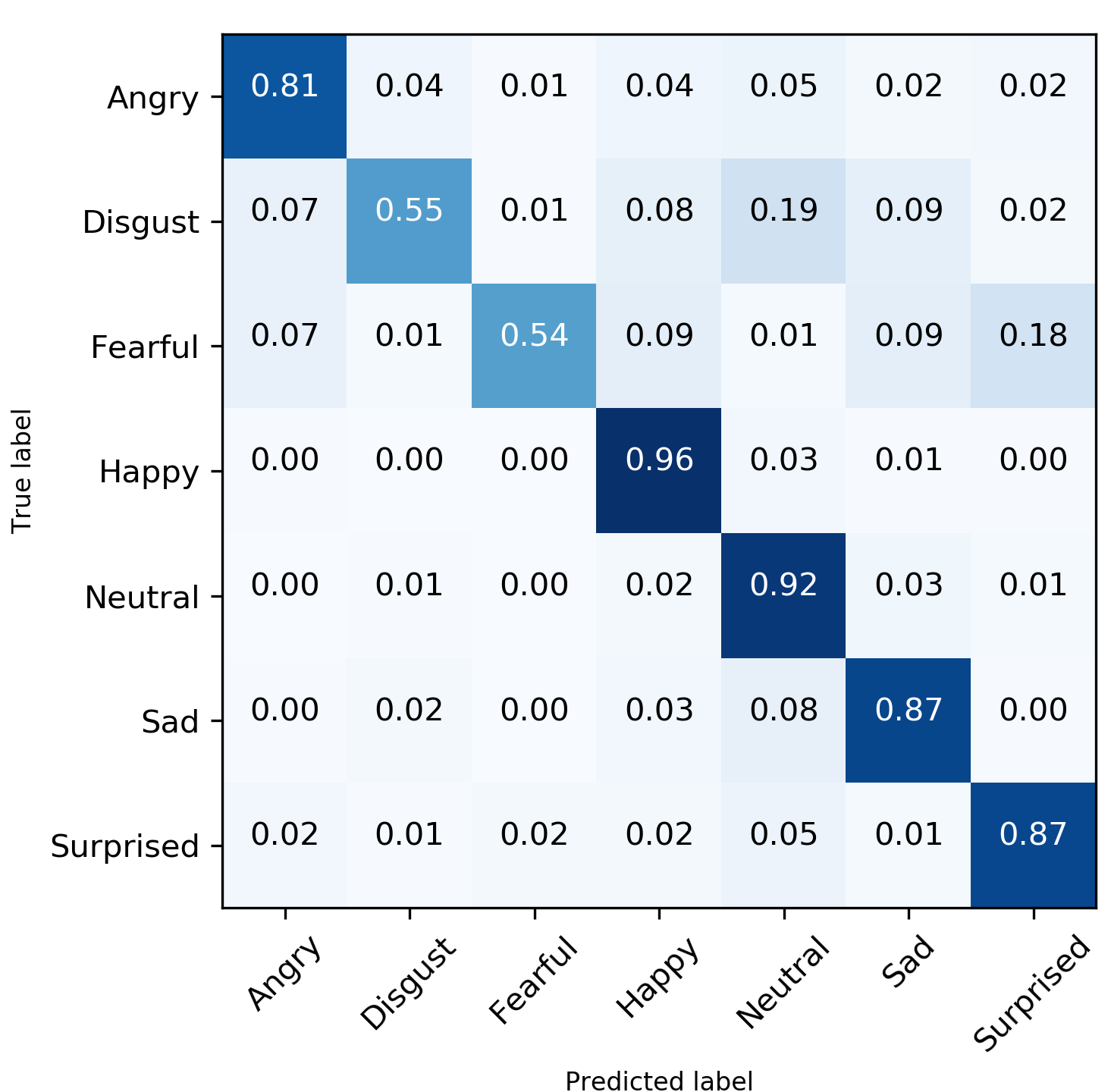}} 
    \subfigure[PT(10\% noise)]{\includegraphics[width=0.3\hsize]{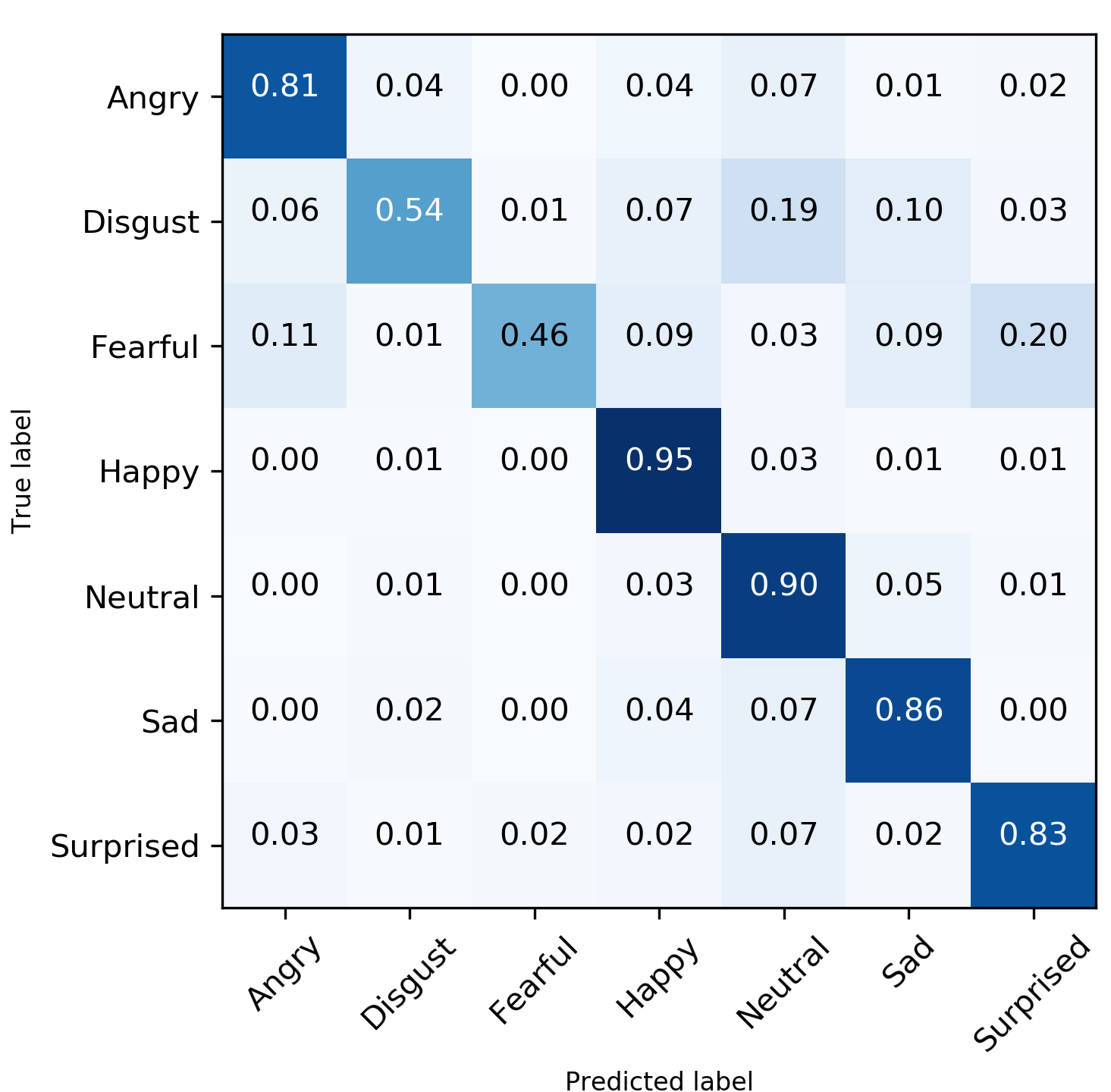}} 
    \subfigure[PT(20\% noise)]{\includegraphics[width=0.3\hsize]{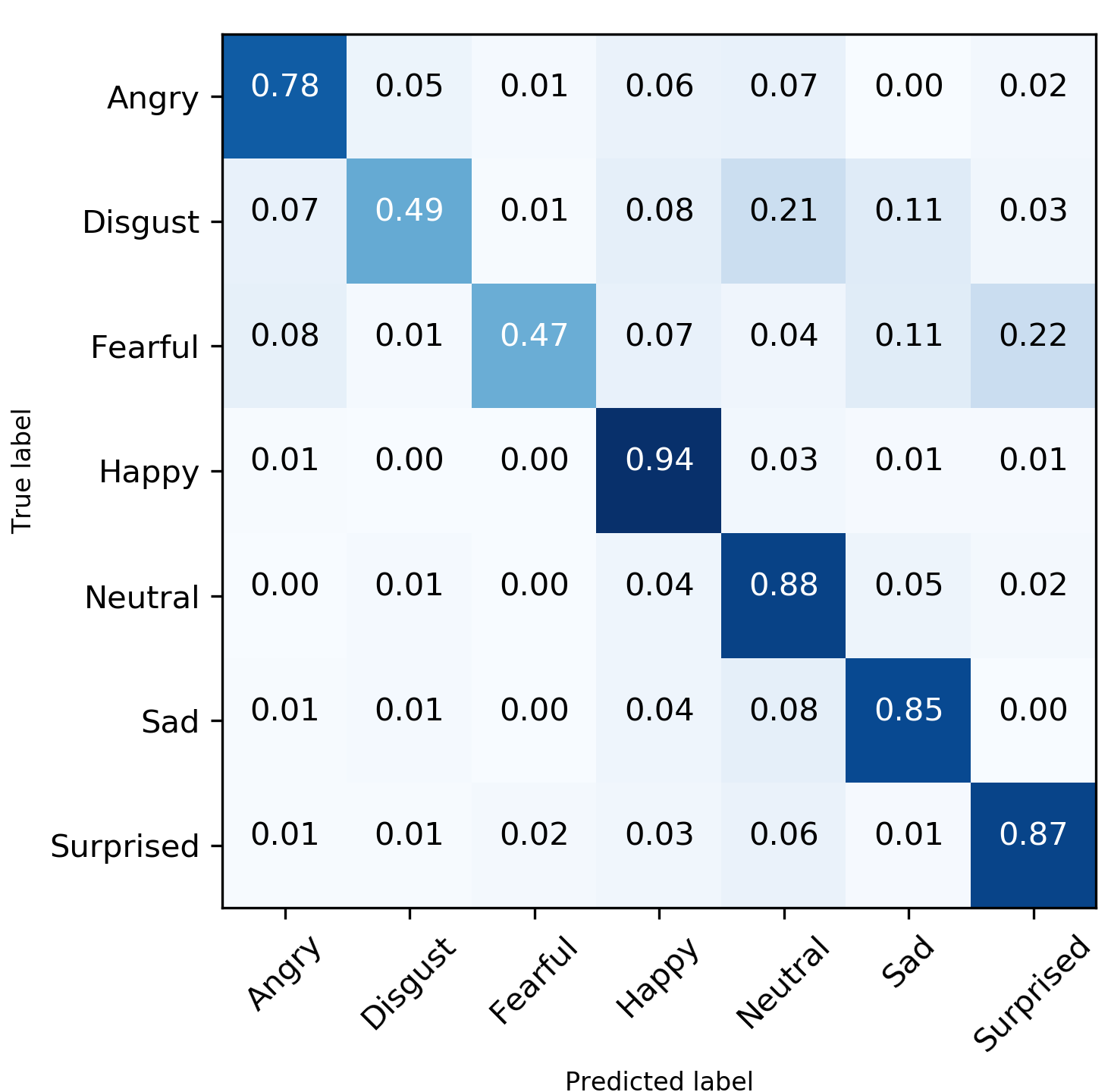}} 
    \subfigure[PT(30\% noise)]{\includegraphics[width=0.3\hsize]{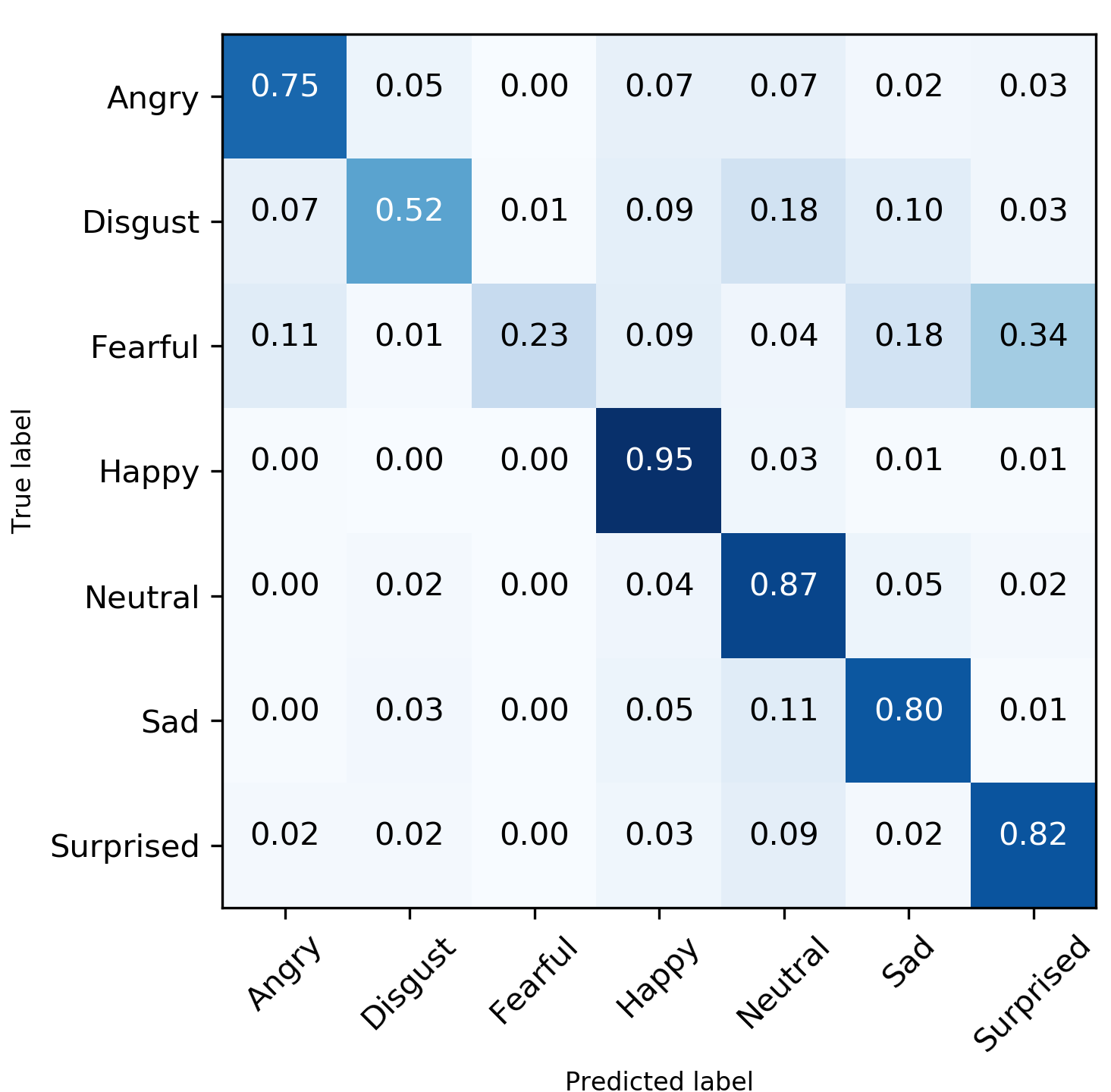}}  
    \caption{\label{confusion_RAF}The confusion matrices on RAF-DB. The top row represents results of three methods on original RAF-DB and the second row represents the results of Progressive Teacher under 3 levels of label noises on RAF-DB test set.}
\end{figure*}
\subsection{Results}
\subsubsection{Evaluations on reliable FER databases.}
    To demonstrate the effectiveness of this semi-supervised approach, we compare our algorithm with five other methods. The fine-tuning baseline means refining the pre-trained network on target dataset (RAF-DB or FERPlus). MixTrain uses all images and their labels in target dataset as well as the auxiliary dataset for training. In PseudoLabel method, we first train a network using target dataset, then classify samples in AffectNet to relabel them. These generated pseudo labels and corresponding images are thereby used to enlarge the target training set. It's worth mentioning that PseudoLabel is a simple semi-supervised method because it doesn't use the original labels of auxiliary data. State-of-art-art methods in semi-supervised learning, Mean Teacher\cite{tarvainen2017mean} and FixMatch\cite{sohn2020fixmatch}, are also provided for comparison. The results are shown in Table \ref{table1}.
        
    \begin{table}[htbp]
        \centering
        \caption{\label{table1}Compare our method with baselines and semi-supervised learning method. ResNet-18 is used as backbone. \dag AffectNet is used as auxiliary dataset. \ddag For RAF-DB, FERPlus is used as auxiliary dataset and vice versa.}
        \begin{tabular}{lcl}
        \toprule
        Method & RAF-DB & FERPlus \\
        \midrule
        Fine-tune(baseline) & 87.09\% & 86.06\% \\
        MixTrain\dag & 85.69\% & 85.40\% \\
        MixTrain\ddag & 87.32\% & 85.85\% \\
        PseudoLabel & 87.39\% & 85.13\% \\
        Mean Teacher\cite{tarvainen2017mean} & 88.41\% & 86.15\% \\
        FixMatch\cite{sohn2020fixmatch} & 87.74\% & 86.45\% \\
        PT(ours) & \textbf{88.69\%} & \textbf{86.60\%} \\
        \bottomrule
        \end{tabular}
    \end{table}
    Observing that the results of MixTrain\dag  are lower than baseline even with the huge amount of labeled images in AffectNet, this may be caused by two reasons. One is the annotation bias in different datasets, which is in accordance with \cite{zeng2018facial}, and the other is the existence of noisy labels in AffectNet. When utilizing two clean datasets which have rare noisy labels, the results of MixTrain\ddag  also declare that enlarging data amount by merging multiple datasets can't improve remarkably and even degrade the performance due to inconsistent annotations. Compared to baseline, PseudoLabel improves the accuracy by 0.3\% on RAF-DB but drops by 0.73\% on FERPlus. The generated pseudo labels are not reliable enough due to inaccurate classifiers and domain shift. Images in RAF-DB shares more similarity with AffectNet so that the performance gets better due to the increase of training data. However, the large domain shift makes this semi-supervised method not work on FERPlus. Without utilizing label information of AffectNet, our Progressive Teacher improves the fine-tuning baselines by 1.6\% on RAF-DB, and 0.54\% on FERPlus respectively. Similar to the results of PseudoLabel, greater data similarity makes greater improvement on RAF-DB. Table \ref{table2} shows state-of-the-art results on RAF-DB.
    \begin{table}[htbp]
        \centering
        \caption{\label{table2}Comparison to the state-of-the-art results on RAF-DB.}
        \begin{tabular}{lcl}
        \toprule
        Method & Accuracy \\
        \midrule
        PAT-ResNet-(gender, race)\cite{cai2018probabilistic} & 84.19\% \\
        IPA2LT(LTNet)\cite{zeng2018facial}& 86.77\%\\
        Acharya et al.\cite{acharya2018covariance} &  87.0\%\\
        Gan et al.\cite{gan2019facial} & 86.31\% \\
        APM-VGG\cite{li2019pooling} & 85.17\% \\
        SCN\cite{wang2020suppressing} & 88.14\% \\
        DAS + VGG-F\cite{liu2020omni}& 86.55\% \\
        SCAN-CCI\cite{gera2021landmark} & 89.02\% \\
        PT-ResNet18 (ours) & 88.69\% \\
        PT-SEResNet50-IR (ours) & \textbf{89.57\%} \\
        \bottomrule
        \end{tabular}
    \end{table}
    Compared with SCN, we both use pre-trained ResNet-18 as backbone and AffectNet as auxiliary data, but we don't employ its label information. This indicates the effectiveness of our semi-supervised approach. Using SEResNet50-IR as backbone, we achieve the best accuracy of 89.57\% on RAF-DB.
\subsubsection{Evaluations on synthetic noisy FER databases.}
\textbf{Uniform Noise}
To correspond with SCN\cite{wang2020suppressing}, which also considers the situation of label noise, we randomly choose 10\%, 20\%, and 30\% of training data for each category and randomly change their labels to others.

\begin{figure}[h]
        \centering
        \includegraphics[width=3.5in]{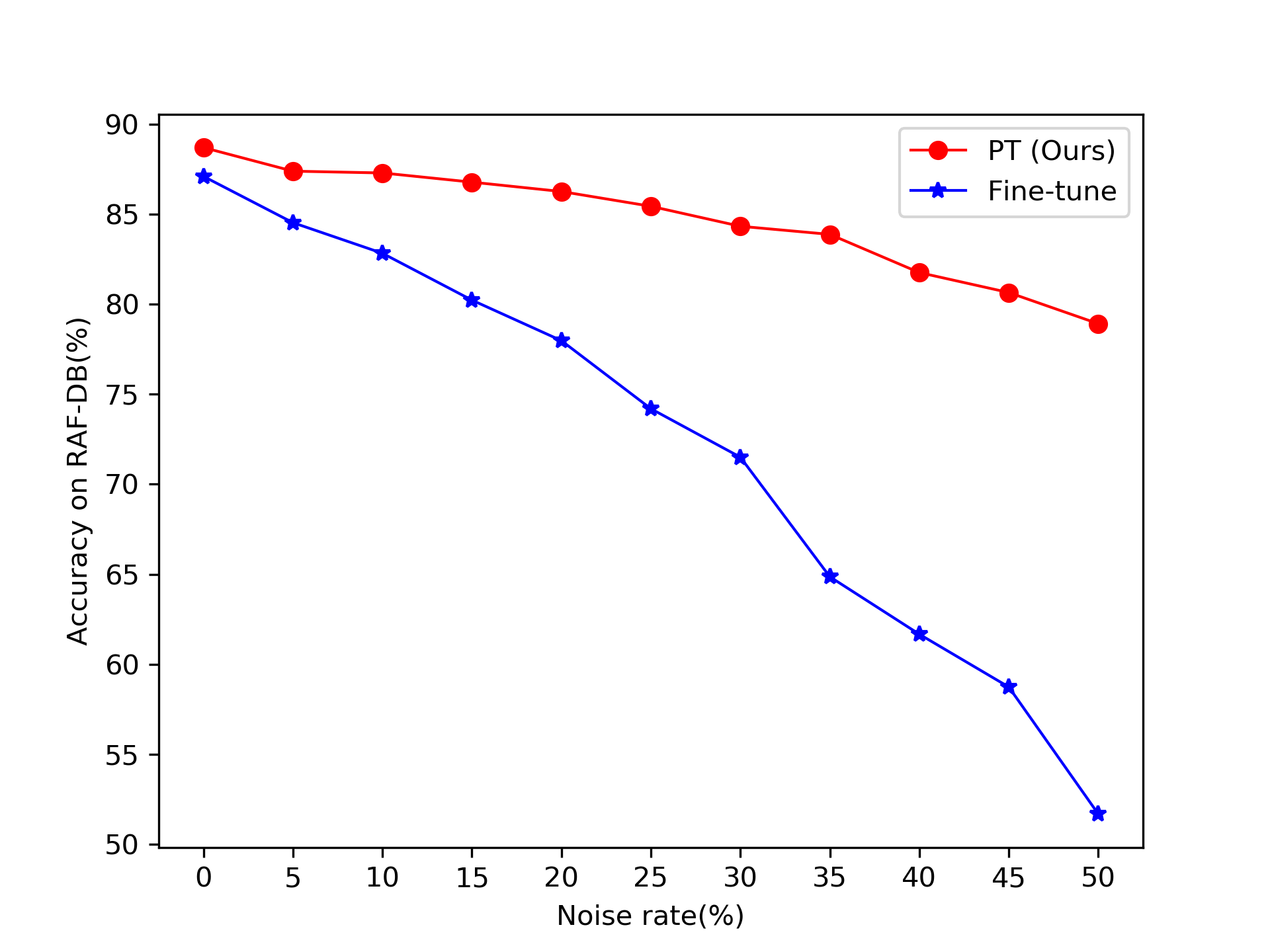}
        \caption{The performance on RAF-DB under detailed noise levels.}
        \label{performance_trend}
    \end{figure}

\begin{table}[htbp]
    \centering
    \caption{\label{table3}The evaluation of PT on synthetic noisy RAF-DB and FERPlus. In SCN, "x" represent fine-tuning the pre-trained ResNet-18, "\checkmark" means SCN algorithm is used. SCN doesn't use auxiliary datasets. \dag Pretrained SEResNet50-IR is used as backbone.}
    \begin{tabular}{ccccc}
        \hline
        Method              & Noise(\%)& RAF-DB  & FERPlus\\ 
        \hline 
        SCN(\xmark)              & 0        & 84.20\% & 86.80\%\\
        SCN(\checkmark)\cite{wang2020suppressing}             & 0        & 87.03\% & 88.01\%\\
        RW Loss\cite{fan2020learning} & 0        & 87.97\% & 87.60\%\\
        SCAN-CCI\cite{gera2021landmark}           & 0        & 89.02\% & 82.35\%\\
        Mean Teacher\cite{tarvainen2017mean}        & 0        & 88.41\% & 86.15\%\\
        \hline
        Fine-tune(baseline) & 0        & 87.09\% & 86.06\%\\
        PT(ours)            & 0        & 88.69\%/$\textbf{89.57\%}^{\dag}$ & 86.60\%\\
        \hline 
        SCN(\xmark)              & 10        & 80.81\% & 83.39\%\\
        SCN(\checkmark)\cite{wang2020suppressing}              & 10        & 82.18\% & 84.28\%\\
        RW Loss\cite{fan2020learning} & 10        & 82.43\% & 83.93\%\\
        SCAN-CCI\cite{gera2021landmark}             & 10        & 84.09\% & 79.25\%\\
        Mean Teacher\cite{tarvainen2017mean}        & 10        & 84.68\% & 82.87\%\\
        \hline
        Fine-tune(baseline) & 10        & 82.82\% & 82.09\%\\
        PT(ours)            & 10        & \textbf{87.28\%} & \textbf{85.07\%}\\
        \hline 
        SCN(\xmark)              & 20        & 78.18\% & 82.24\%\\
        SCN(\checkmark)\cite{wang2020suppressing}             & 20        & 80.80\% & 83.17\%\\
        RW Loss\cite{fan2020learning} & 20        & 80.41\% & 83.55\%\\
        SCAN-CCI\cite{gera2021landmark}             & 20        & 78.72\% & 72.93\%\\
        Mean Teacher\cite{tarvainen2017mean}        & 20        & 81.40\% & 82.87\%\\
        \hline
        Fine-tune(baseline) & 20        & 77.97\% & 77.97\%\\
        PT(ours)            & 20        & \textbf{86.25\%} & \textbf{84.27\%}\\
        \hline 
        SCN(\xmark)              & 30        & 75.26\% & 79.34\%\\
        SCN(\checkmark)\cite{wang2020suppressing}              & 30        & 77.46\% & 82.47\%\\
        RW Loss\cite{fan2020learning} & 30        & 76.77\% & 82.75\%\\
        SCAN-CCI\cite{gera2021landmark}             & 30        & 70.99\% & 68.90\%\\
        Mean Teacher\cite{tarvainen2017mean}       & 30        & 75.00\% & 72.06\%\\
        \hline
        Fine-tune(baseline) & 30        & 71.50\% & 70.03\%\\
        PT(ours)            & 30        & \textbf{84.32\%} & \textbf{83.73\%}\\
        \hline 
    \end{tabular}
\end{table}
    In Table \ref{table3}, we evaluate our PT algorithm under different levels of label noises on RAF-DB and FERPlus to demonstrate its effectiveness. We report the result when the network becomes stable instead of selecting the highest accuracy during training process. The results of state-of-the-art supervised methods on FER\cite{wang2020suppressing,fan2020learning,gera2021landmark} under different levels of noise are also shown to make comparison. The backbones of SCN\cite{wang2020suppressing} and SCAN-CCI\cite{gera2021landmark} are ResNet-18 and ResNet-50, respectively. We can see that on original RAF-DB, Progressive Teacher surpasses Mean Teacher by only 0.28\%, because the dataset contains rare noise. Under the noise rate of 10\%, 20\% and 30\% on RAF-DB, our method all achieves the highest accuracy and improves the baseline by 4.46\%, 8.28\% and 12.82\% respectively. The effect of our PT algorithm gets more remarkable as the noisy labels increase. Results on FERPlus show the same regularity. When the noise rate variates from 10\% to 30\%, the accuracy improvements are 2.98\%, 6.30\% and 13.7\%. Though we get inferior accuracy on original FERPlus due to the domain discrepancy between FERPlus and AffectNet, we obtain superior performance on synthetic noisy datasets which indicates the advantage of semi-supervised approach and sample selection. Additionally, even adding 30\% noises to original RAF-DB and FERPlus, the performance drops by only 4.37\% and 2.87\% respectively with the help of PT while SCN degrades by 9.57\% and 5.54\%. This indicates that our method can work in extremely hard condition. To see the performance's changing trend, we add more detailed noise levels on RAF-DB and the result is shown in Fig.\ref{performance_trend}.
    
    Fig.\ref{confusion_RAF} shows the confusion matrices on RAF-DB. We can find that disgust and fearful are the hardest two expressions to recognize, which can be explained that they contain least samples. The class imbalance problem leads to the variance of recognition performance between different expressions. Our Progressive Teacher improves the overall accuracy but degrades on difficult expressions such as disgust and fearful. This is the natural consequence of data selection since samples in small-class are relatively hard to recognize and will have larger loss values so that some of them are likely to be filtered out by network. We focus on this problem and extend our method in section 4.6.

\begin{figure}[htbp]
        \centering
        \subfigure[Performances on RAF-DB under asymmetric noise.]{\includegraphics[width=0.98\hsize]{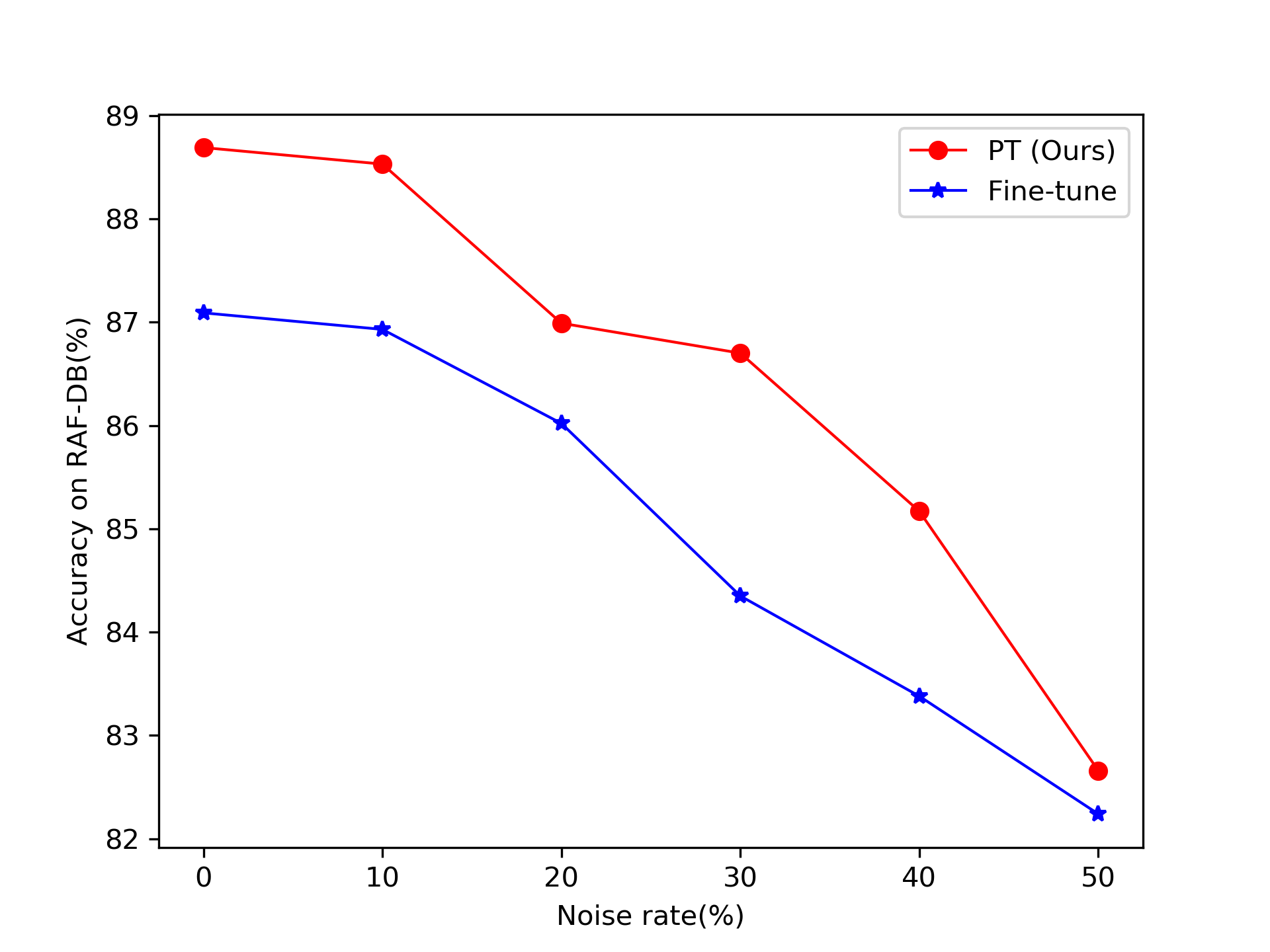}} 
        \subfigure[Performances on FERPlus under asymmetric noise.]{\includegraphics[width=0.98\hsize]{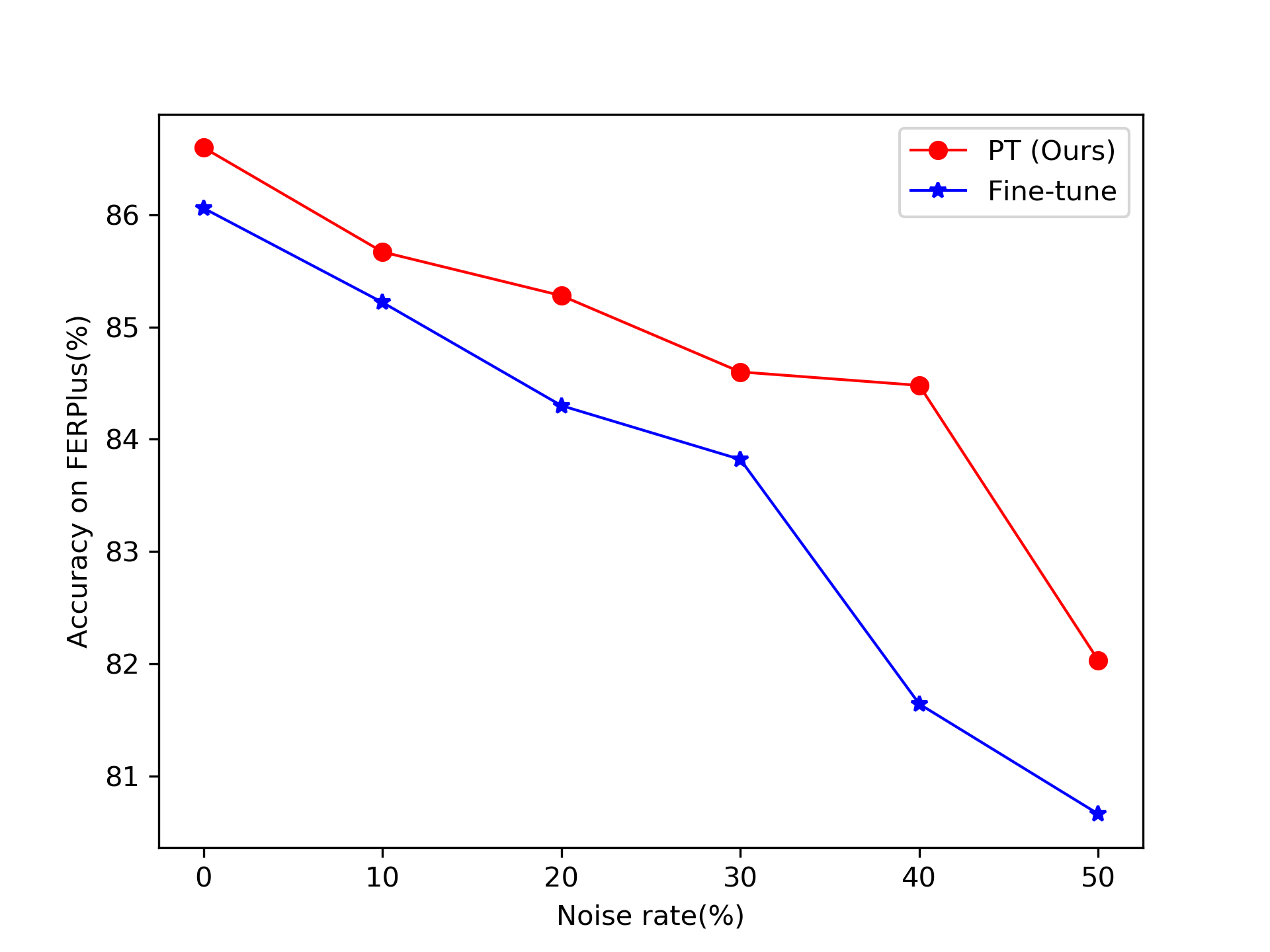}}  
        \caption{\label{asymmetric}The performance of Progressive Teacher when adding asymmetric noise to RAF-DB and FERPlus. In this scenario, fearful and surprised are relabeled to each other.}
    \end{figure}

    \textbf{Asymmetric Noise}
    In real situation, expression surprised is easy to be annotated as expression fearful and vice versa, we also conduct experiments under asymmetric noise. Specifically, samples in class fearful and class surprised are annotated to each other in the same ratio. The experimental result is recorded in Fig. \ref{asymmetric}. We can see that our method also achieves better performance than fine-tuning baseline.
\subsubsection{Evaluations on real-world noisy FER database AffectNet.}
\begin{figure}[htbp]
        \centering
        \subfigure[Fine-tune]{\includegraphics[width=0.95\hsize]{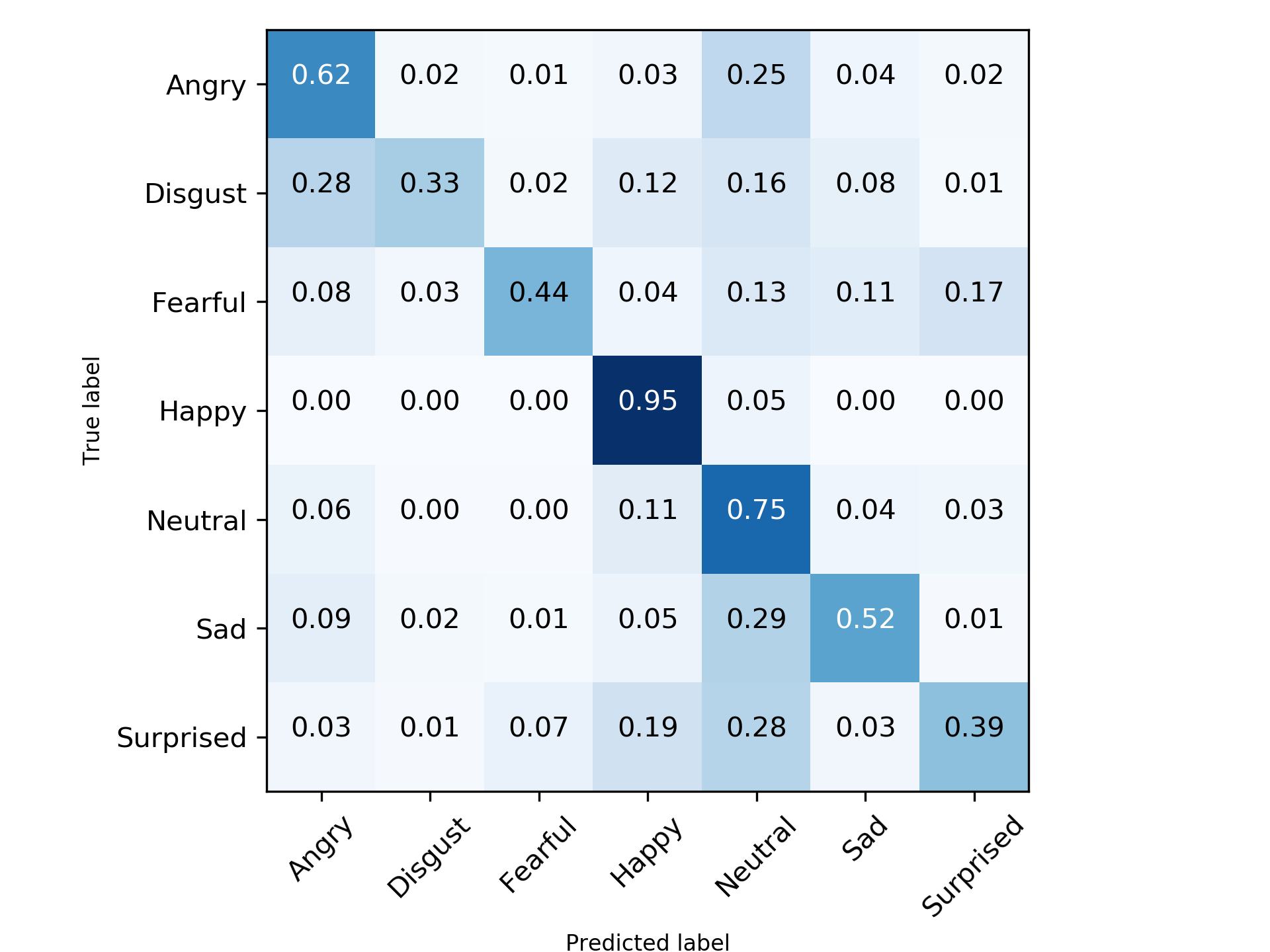}} 
        \subfigure[Mean Teacher]{\includegraphics[width=0.95\hsize]{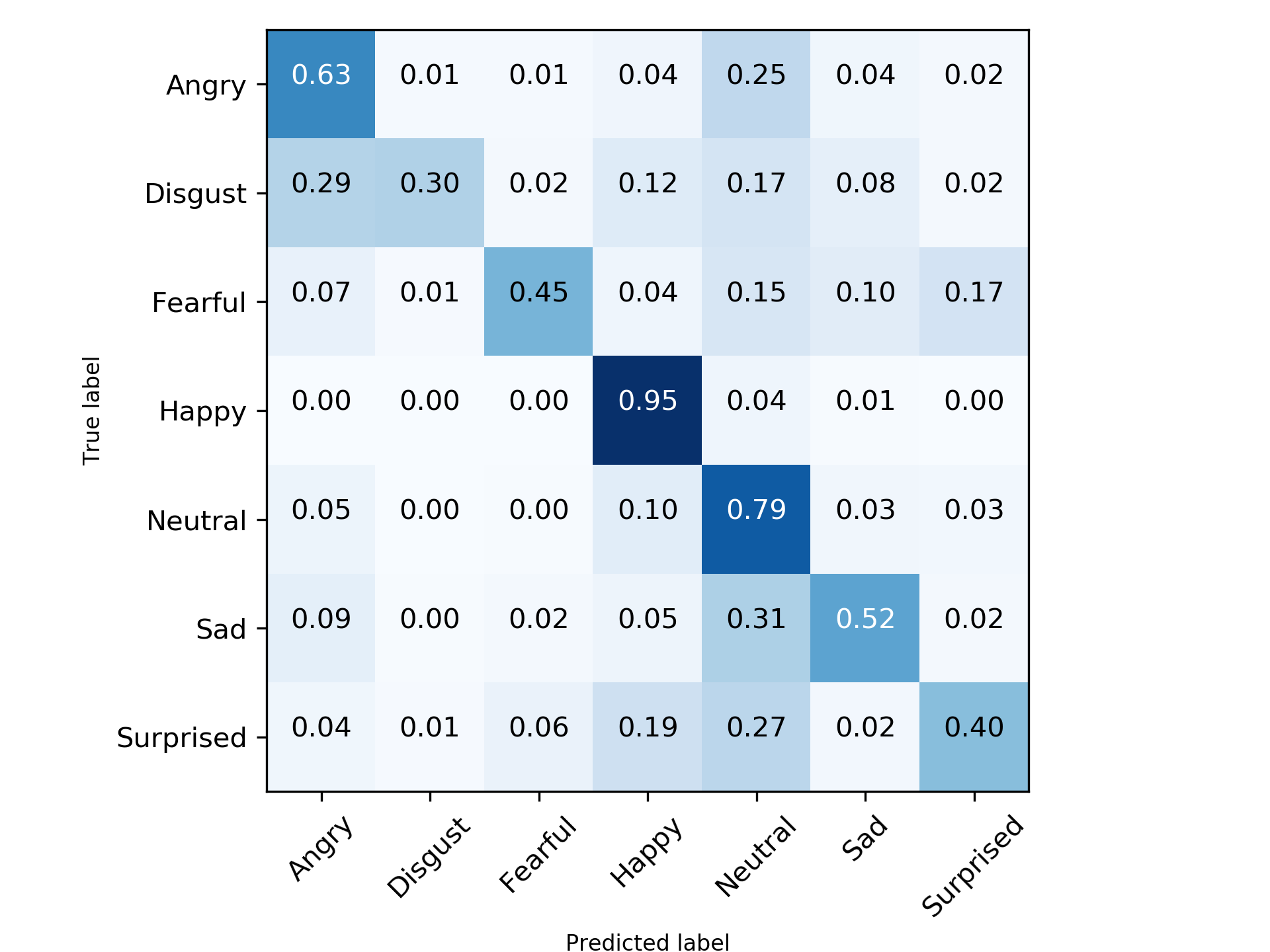}}
        \subfigure[Progressive Teacher]{\includegraphics[width=0.95\hsize]{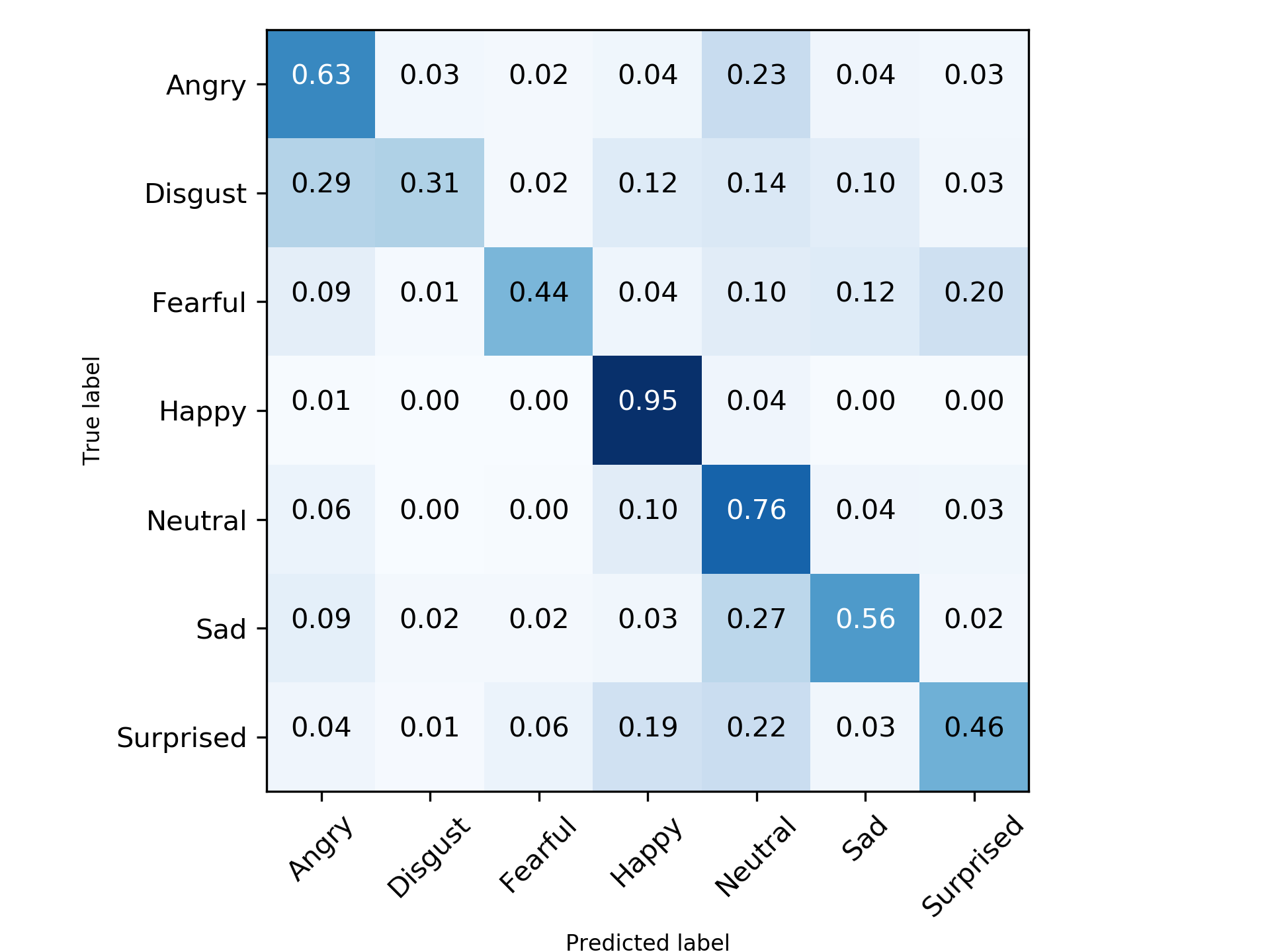}} 
        \caption{\label{confusion_Aff}The confusion matrices on AffectNet.}
    \end{figure}

    We also conduct experiments on AffectNet, a real noisy database. FERPlus dataset is utilized as unlabeled data. The size of mini-batch is set to 200 and half of them are labeled data. The abandoning rate r is set to 0.2, T is set to 3000. Besides, among selected large-loss samples, disgust, fearful and surprised are reserved with the probability of 0.9, angry and sad are reserved with the probability of 0.8. We compare our method with fine-tuning baseline and Mean Teacher in Table. \ref{Acc_AffectNet}.
    \begin{table}[htbp]
        \centering
        \caption{The accuracy when training with AffectNet.}
        \begin{tabular}{cccccccc}
        \toprule
        &Test on AffectNet & Test on RAF-DB\\
        \midrule
        Fine-tune&57.23\%&79.04\%\\
        Mean Teacher\cite{tarvainen2017mean}&57.62\%&81.19\%\\
        PT(ours)& 58.54\% &82.11\%\\     
        \bottomrule
        \label{Acc_AffectNet}
        \end{tabular}
    \end{table}
    The confusion matrices on AffectNet test set is shown in Fig.\ref{confusion_Aff}.

\subsection{Visualizations}

\begin{figure}[htbp]
        \centering
        \includegraphics[scale=0.45]{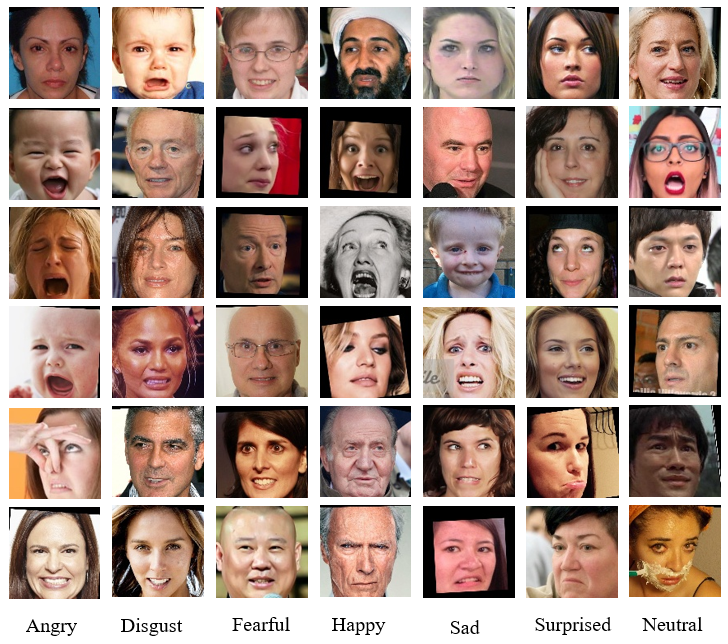}
        \caption{Filtered training samples with Progressive Teacher.}
        \label{picked_example_Affect}
    \end{figure}

    To visualize the effect of PT algorithm, we show some examples which are filtered out as noisy samples in Fig.\ref{picked_example_Affect}. This experiment is conducted on AffectNet. The selected training samples are with the largest loss values in each mini-batch. We also list the data distribution of samples in Table \ref{number}. 
 
\begin{figure}[htbp]
        \centering
        \subfigure[10\%]{\includegraphics[width=0.3\hsize]{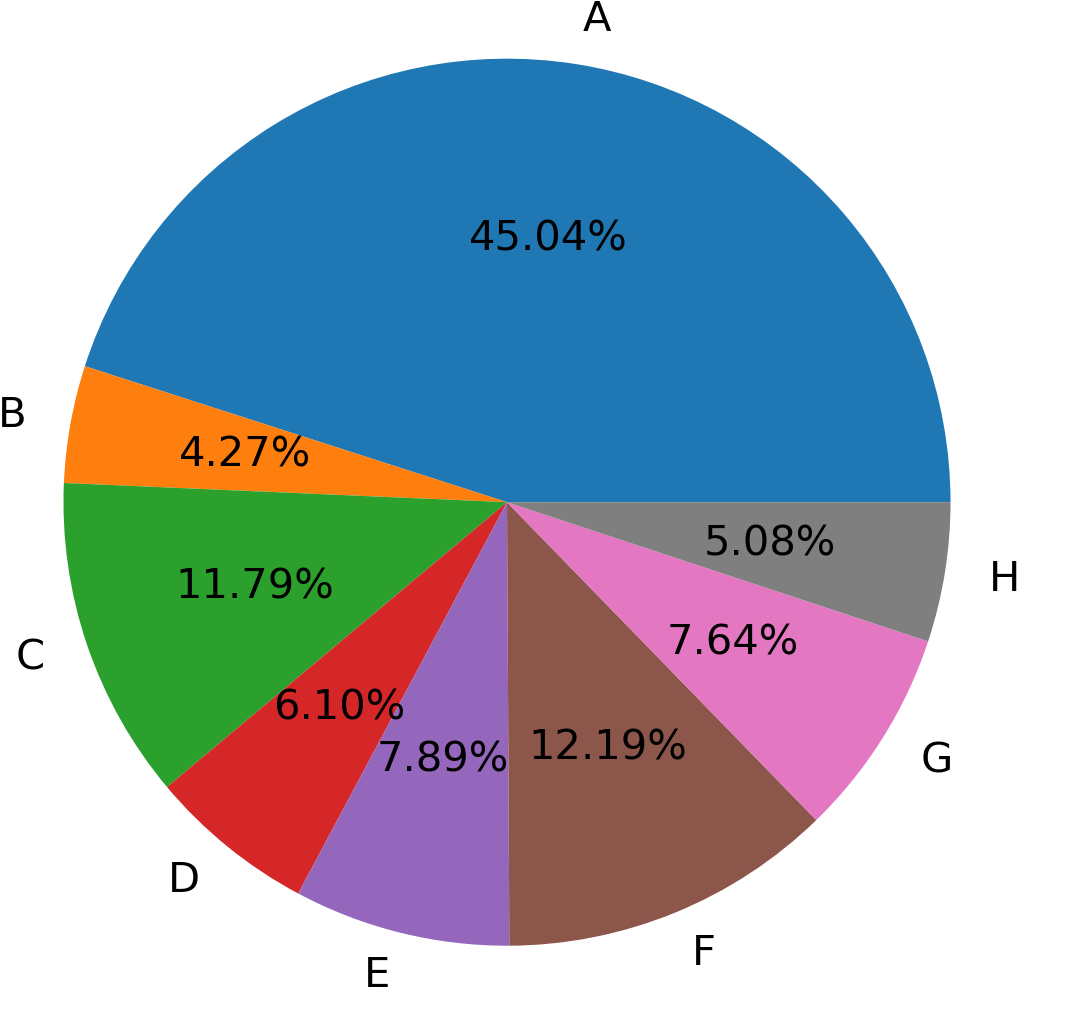}} 
        \subfigure[20\%]{\includegraphics[width=0.3\hsize]{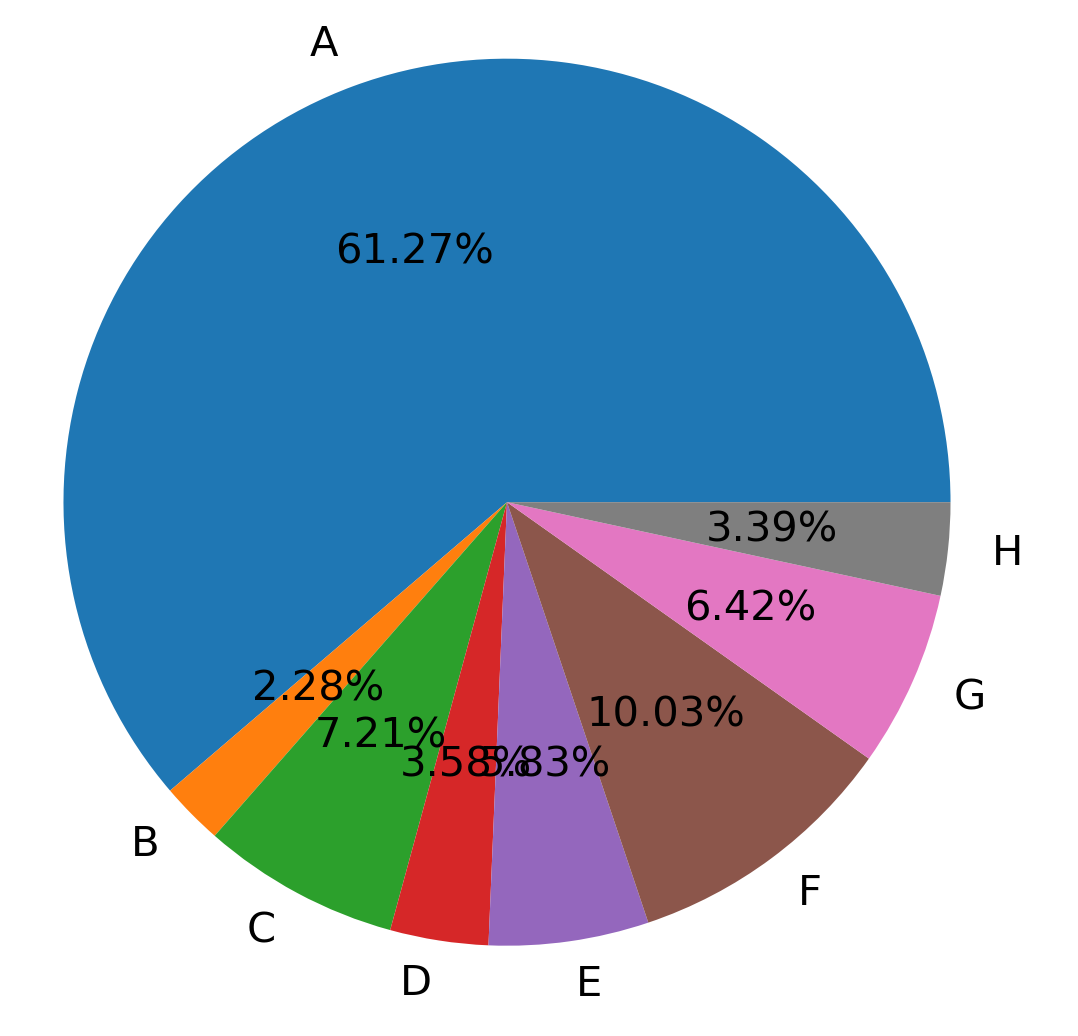}} 
        \subfigure[30\%]{\includegraphics[width=0.3\hsize]{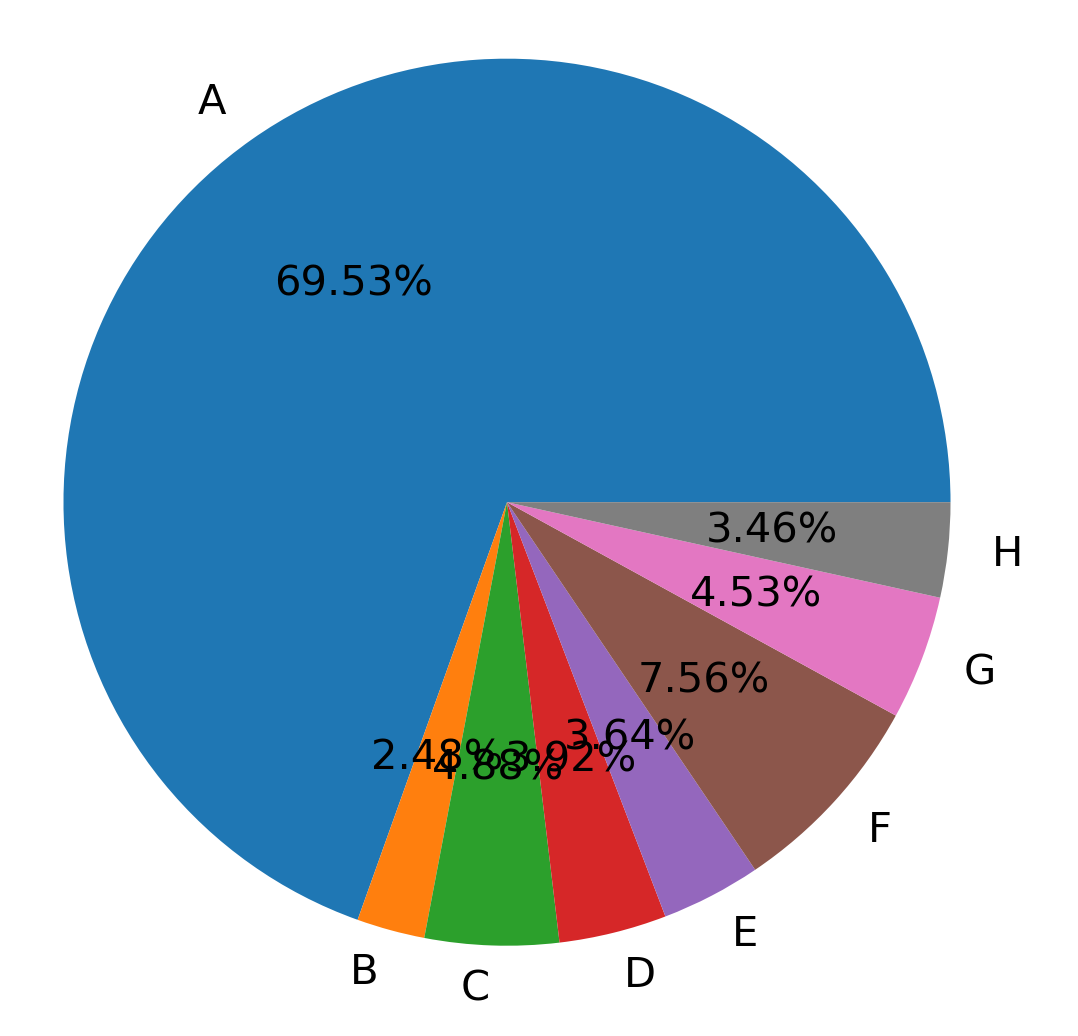}} 
        \caption{\label{bing}The distribution of filtered samples in (A)synthetic noisy labels, (B)Angry, (C)Disgust, (D)Fearful, (E)Happy, (F)Neutral, (G)Sad, (F)Surprised when adding noise to RAF-DB.}
    \end{figure}

    We add synthetic noise to RAF-DB since it contains rare noisy labels. Fig.\ref{bing} and Table \ref{number_2} shows the data distribution of abandoned samples under different noise levels on RAF-DB. Note that we abandon r'+10\% large-loss samples in mini-batch, in which the synthetic noise ratio is r'. The majority of abandoned samples are synthetic noisy labels and the rest are with original label. 
    
    \begin{table}[htbp]
        \centering
        \caption{The number of samples in each expression class of original datasets and filtered out data.}
        \setlength{\tabcolsep}{1.2mm}
        \begin{tabular}{cccccccc}
        \toprule
        &Ang&Dis&Fea&Hap&Neu&Sad&Sur\\
        \midrule
        RAF-DB&705&717&281&4772&2524&1982&1290\\
        Filtered&49&149&96&52&46&63&37\\
        \hline 
        FERPlus&2399&175&648&7410&9365&3403&3378\\
        Filtered&102&72&79&86&347&297&89\\
        \hline 
        AffectNet&24882&3803&6378&134415&74874&25459&14090\\
        Filtered&1548&219&346&9078&19180&1843&592\\
        \bottomrule
        \label{number}
        \end{tabular}
    \end{table}
    \begin{table}[htbp]
        \centering
        \caption{The distribution of filtered out images on synthetic noisy RAF-DB.}
        \setlength{\tabcolsep}{1.2mm}
        \begin{tabular}{ccccccccc}
        \toprule
        Noisy rate&Synthetic Noise&Ang&Dis&Fea&Hap&Neu&Sad&Sur\\
        \midrule
        10\%&1108&105&290&150&194&300&188&125\\
        20\%&2261&84&266&132&215&370&237&125\\
        30\%&3421&122&240&193&179&372&223&170\\
        \bottomrule
        \label{number_2}
        \end{tabular}
    \end{table}
  
\subsection{Ablation Study}
\begin{figure}[htbp]
    \centering
    \includegraphics[scale=0.4]{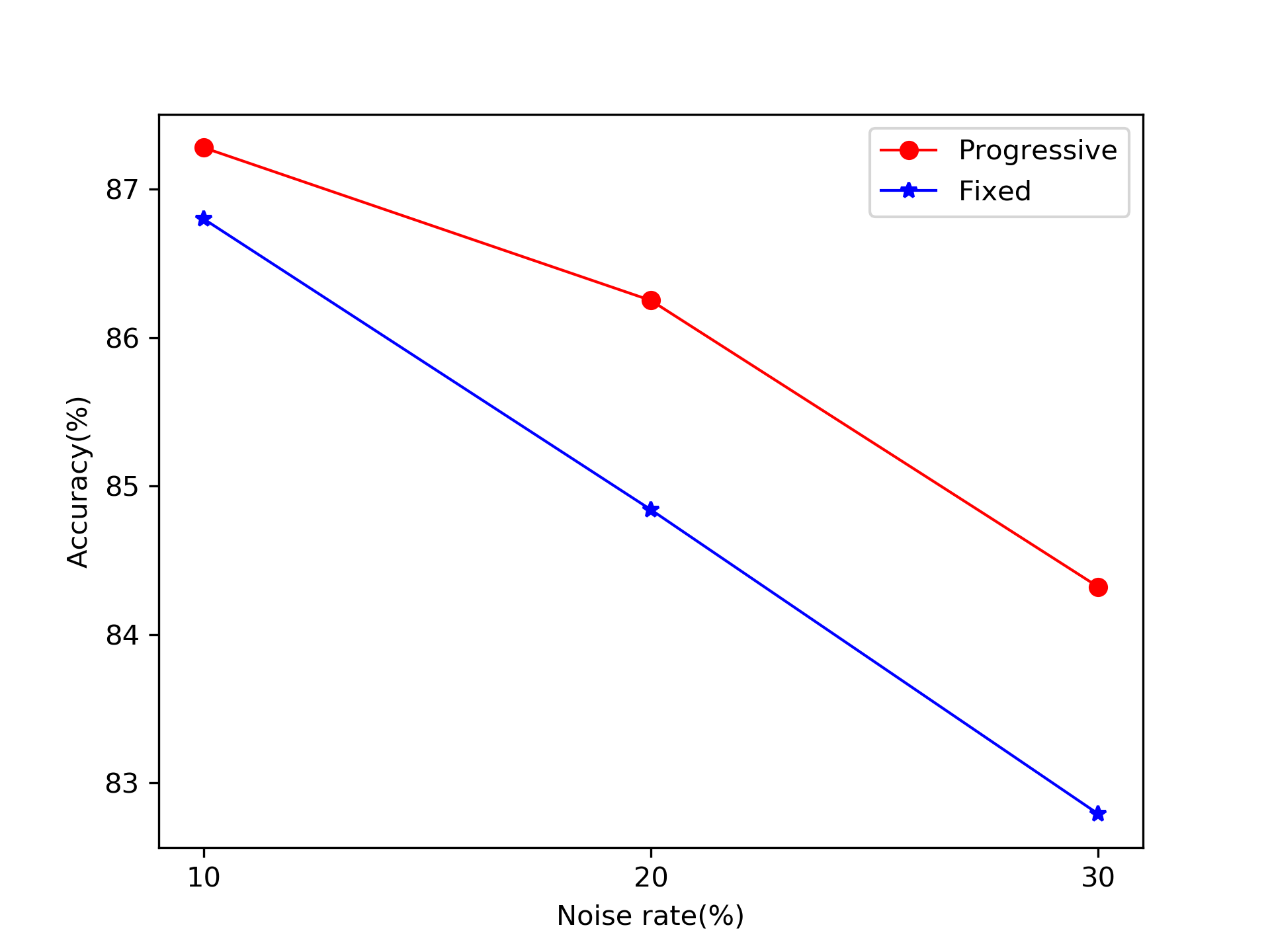}
    \caption{The comparison between abandoning samples progressively and with the same ratio all through. Symmetric noise is adopted.}
    \label{stable_progressive}
\end{figure}

    Progressive Teacher selects samples progressively which means that it utilizes less labeled data as training continues.
    In this subsection, we first conduct experiment to demonstrate the effectiveness of selecting samples progressively instead of keeping the same ratio all through, which is shown in Fig.\ref{stable_progressive}. The blue line represents using the same selecting ratio $R=1-r$. To correspond with the setting in Progressive Teacher, we set r to 0.2, 0.3 and 0.4 respectively as the noise rate increases. We can see that progressive selection outperforms using fixed selecting ratio under three noise rate levels. The noise rate gets higher, the former selection method gets more effective.

\begin{figure}[htbp]
        \centering
        \includegraphics[scale=0.4]{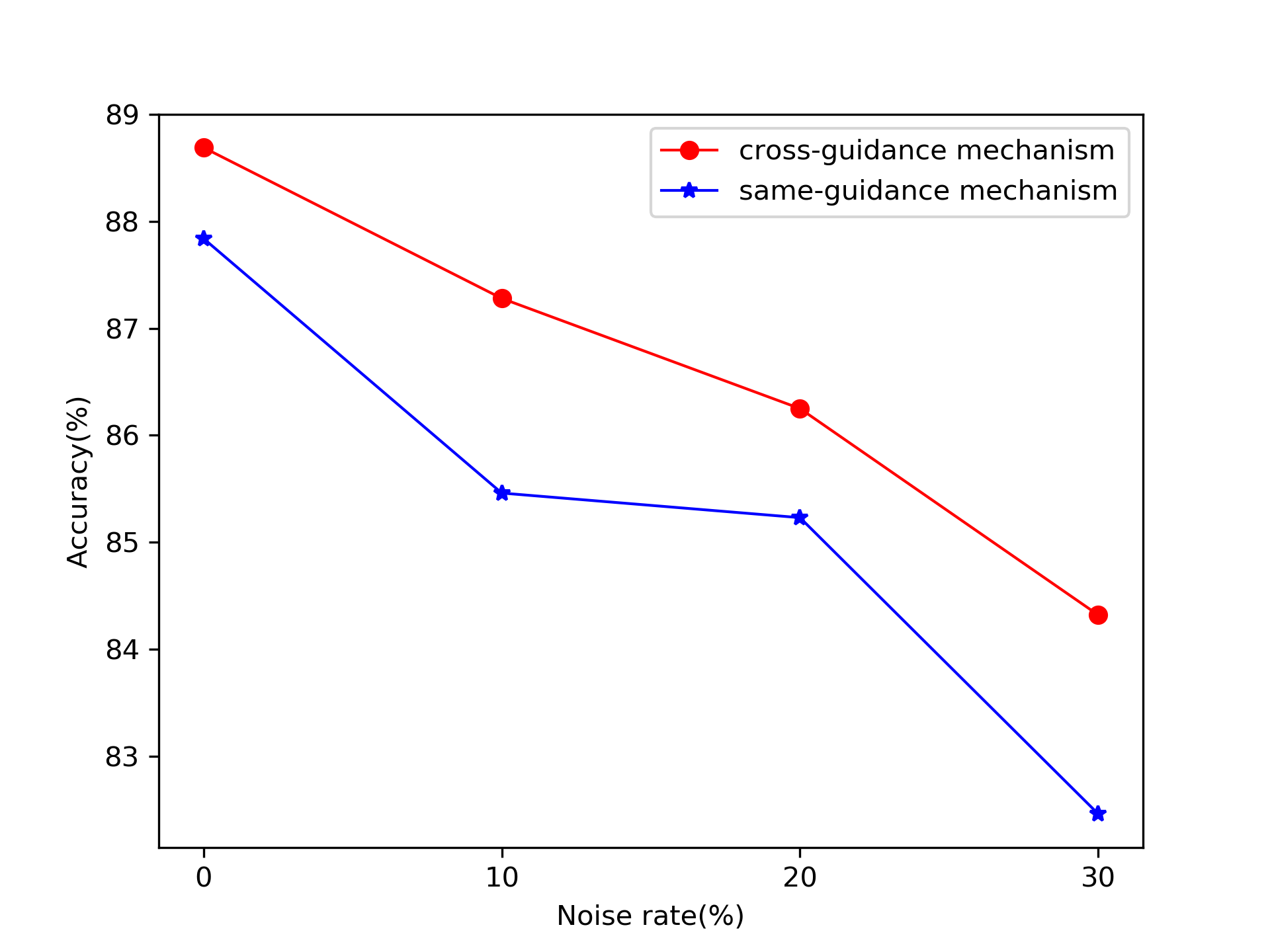}
        \caption{The comparison between same-guidance and cross-guidance mechanism on RAF-DB. Symmetric noise is adopted.}
        \label{same_vs_cross}
    \end{figure}

    We also conduct experiments to illustrate the effectiveness of cross-guidance mechanism over same-guidance. When using one group of teacher-student model, which means that the teacher model feeds the student with clean samples and we compute the consistency loss between them, the accuracy of same-guidance mechanism on RAF-DB is 87.84\%, which is lower than the 88.69\% accuracy in cross-guidance setting by 0.85\%. This indicates that the cross-guidance mechanism is superior to same-guidance. Additionally, considering there are two teacher-student groups but we compute the consistency loss between TNet-1 and SNet-1 (TNet-1 feeds SNet-2 with clean samples), we achieve accuracy of 88.10\% on RAF-DB. When using two pairs of models, they have different weight initializations and thus will feed each other with different samples in the early stage of training. This will facilitate the variance of learning in two student models. And this variance will accumulate in their teachers. Different training samples make the two groups have different learning ability. In this circle, we argue that the two groups complement each other and will learn better. Moreover, adding more groups is difficult to implement and brings more computing cost. Therefore, two pairs of networks is a good choice. We compare the performance of same-guidance and cross-guidance in Fig.\ref{same_vs_cross}.

\subsection{Further Study}
\begin{figure}[htbp]
        \centering
        \includegraphics[scale=0.35]{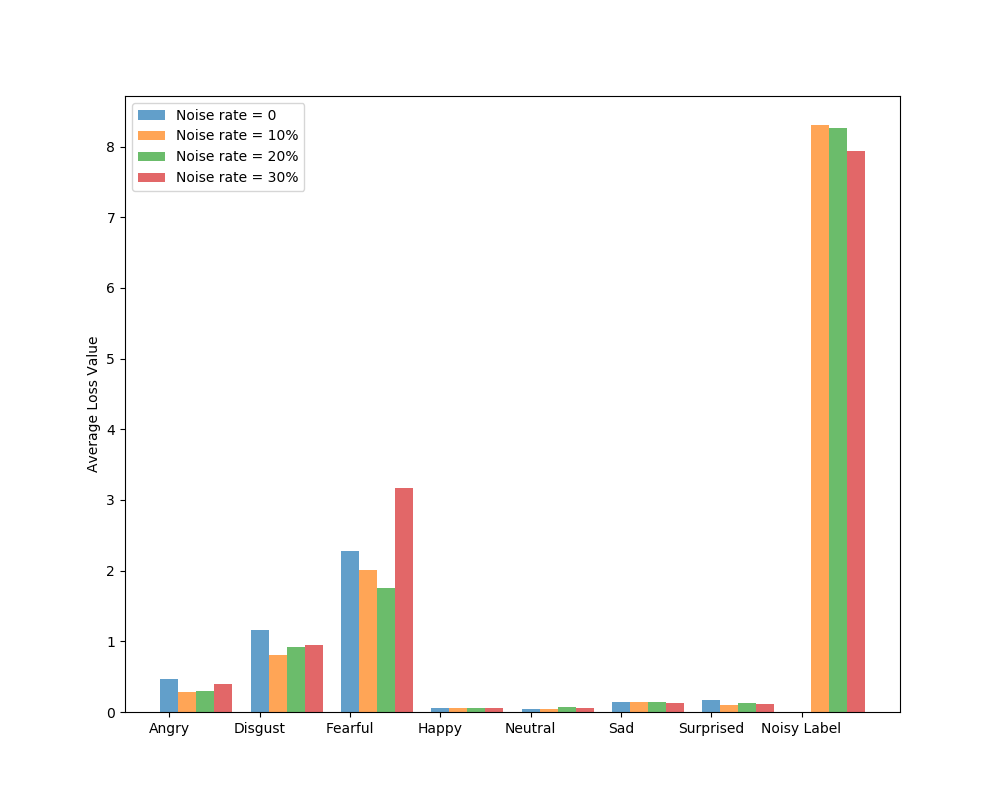}
        \caption{The Average loss value of seven expressions and noisy labels on RAF-DB.}
        \label{loss}
    \end{figure}

\begin{figure}[htbp]
        \centering
        \includegraphics[scale=0.40]{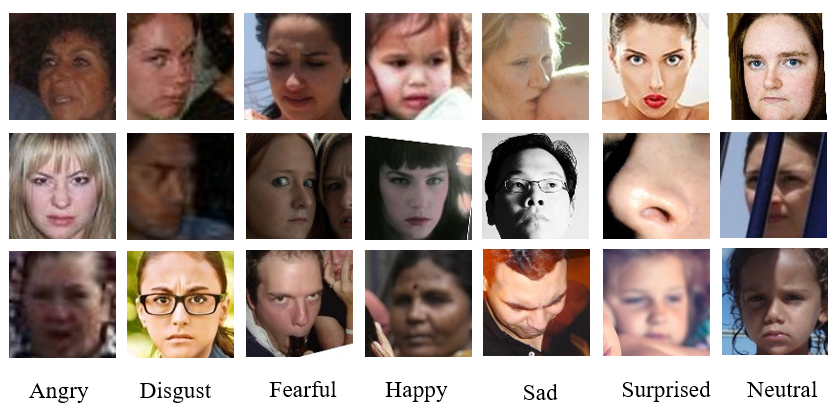}
        \caption{Filtered training samples with Progressive Teacher and confidence estimator on original RAF-DB.}
        \label{filtered_confidence}
    \end{figure}

\begin{figure}[htbp]
        \centering
        \subfigure[No noise (Acc=88.59\%)]{\includegraphics[width=0.48\hsize]{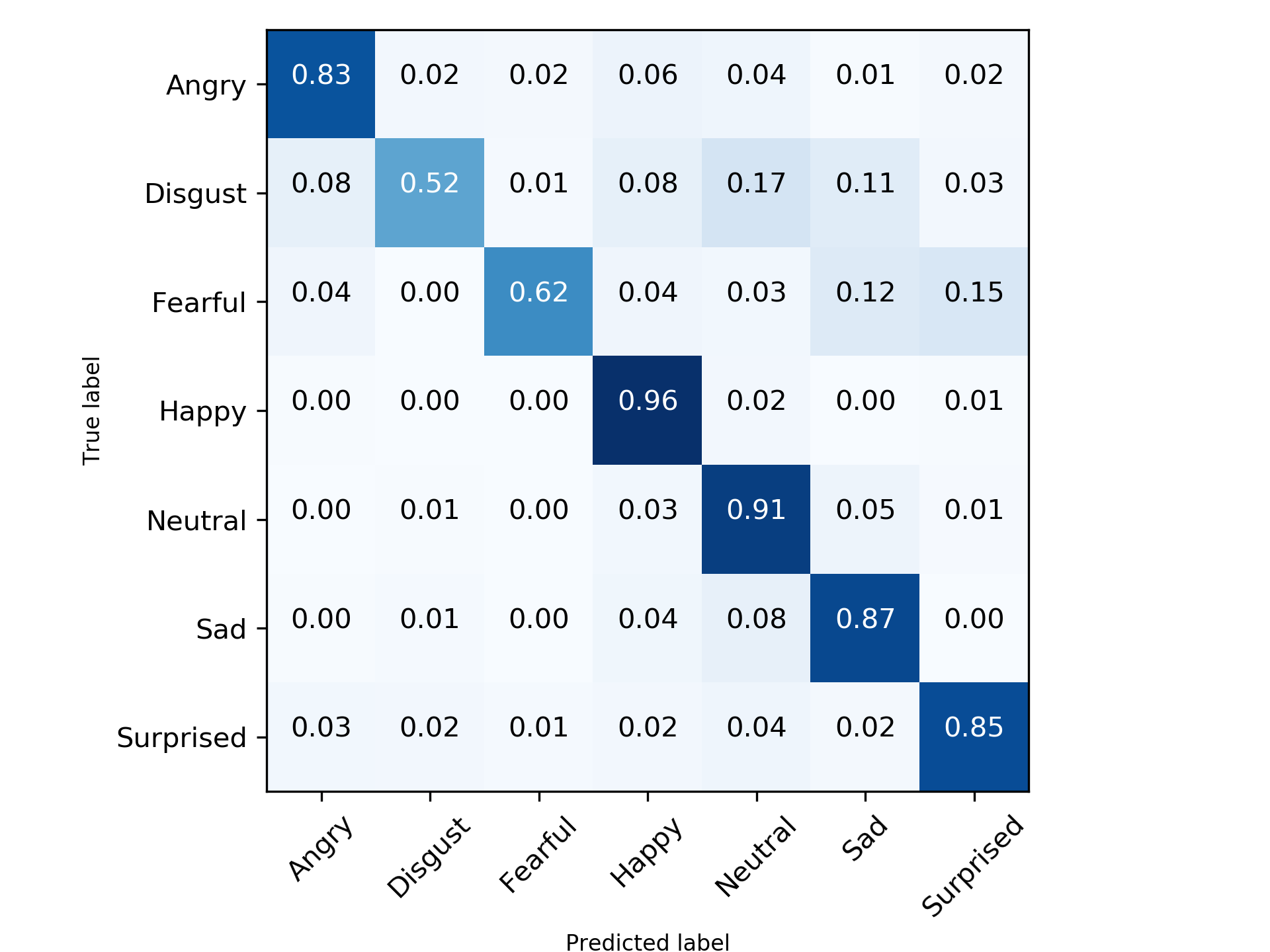}} 
        \subfigure[10\% noise (Acc=87.42\%)]{\includegraphics[width=0.48\hsize]{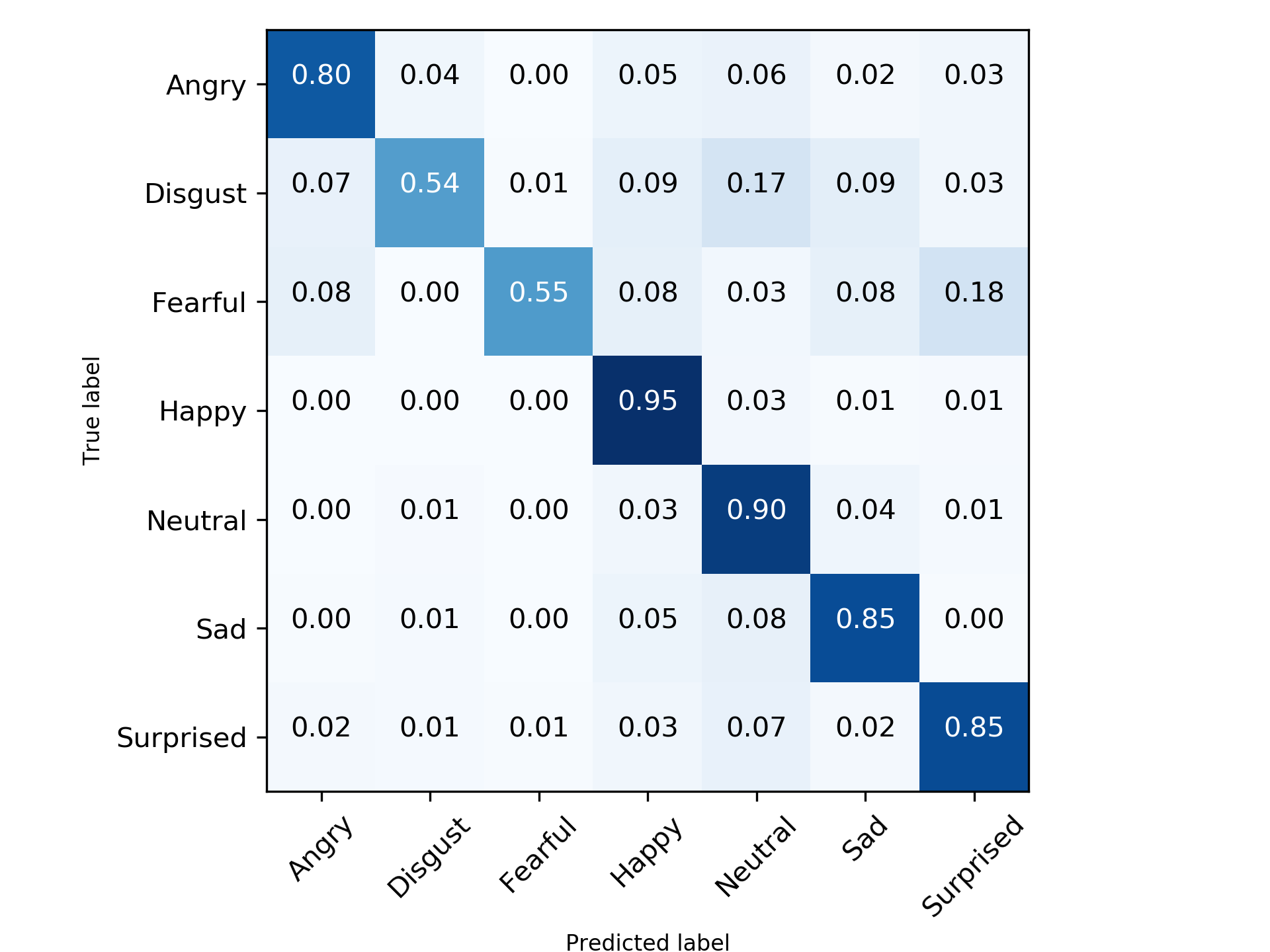}} 
        \subfigure[20\% noise (Acc=86.47\%)]{\includegraphics[width=0.48\hsize]{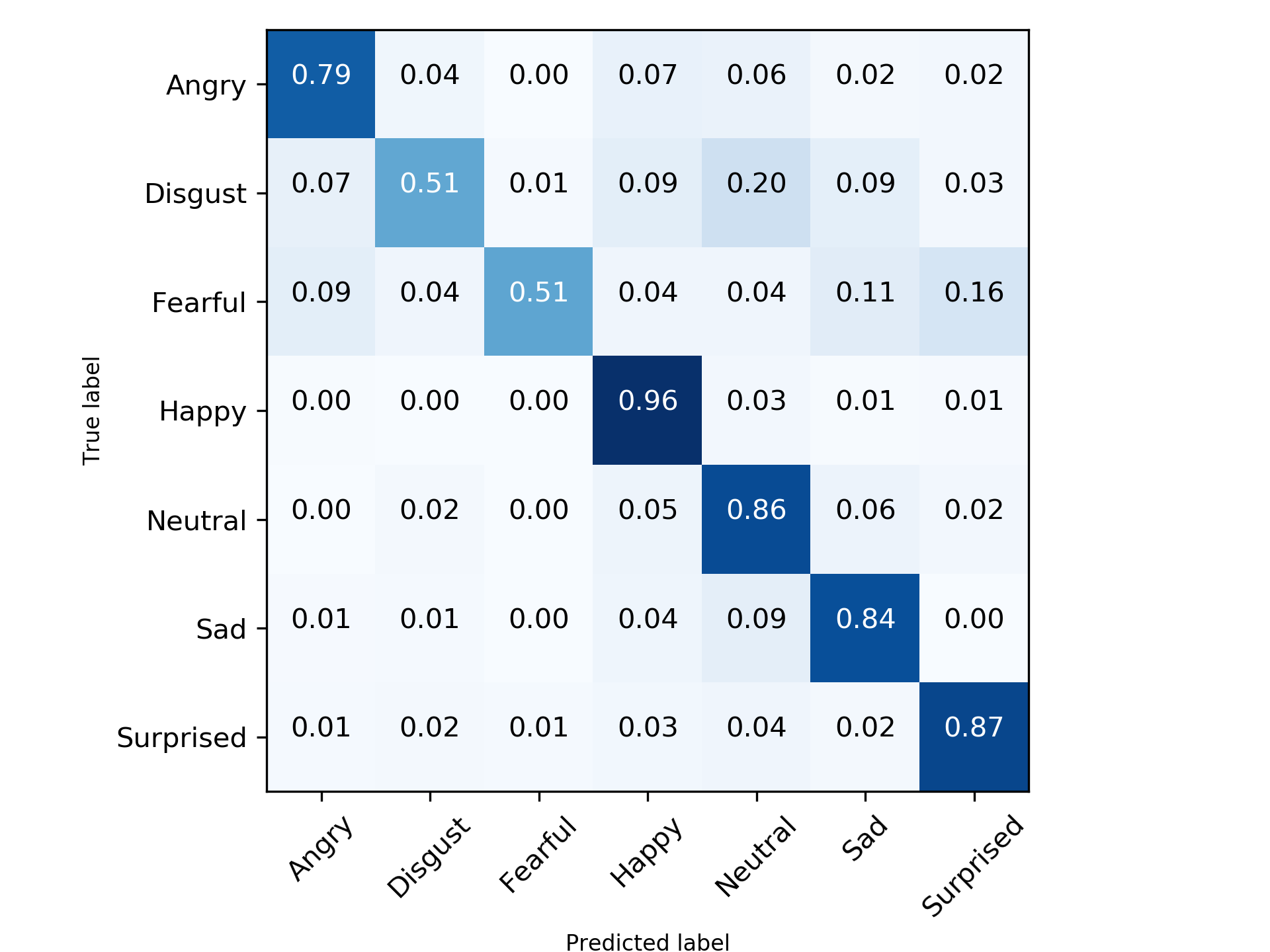}} 
        \subfigure[30\% noise (Acc=84.78\%)]{\includegraphics[width=0.48\hsize]{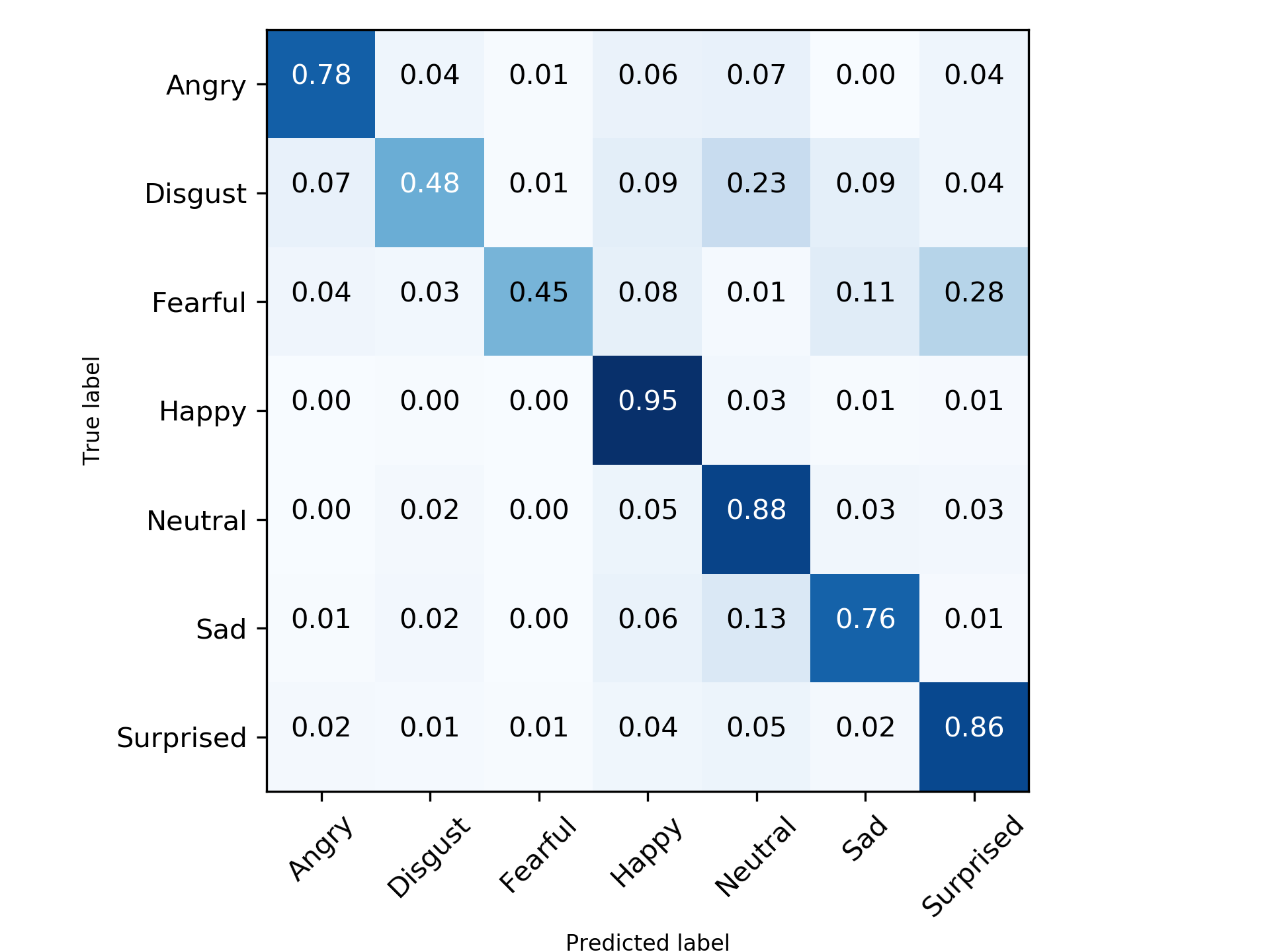}}  
        \caption{\label{confusion_RAF_2}The confusion matrices on RAF-DB when extending PT with confidence estimator.}
    \end{figure}

    In this section, we will analyze the relationship between loss value and noisy label first and then introduce an extension version of our method. Our experiments are conducted on RAF-DB. We count the average loss value of seven expressions and noisy labels and plot them in Fig.\ref{loss}. It can apparently be observed that noisy labels have much larger loss values than others. However, due to the lack of samples in some expressions(i.e., angry, disgust and fearful) and the difficulty of recognition, the loss value of these categories are larger than others (but smaller than noisy labels). In order not to improperly abandon these samples, we guide our Progressive Teacher with a confidence discriminator. Specifically, we train a confidence estimator which learns the confidence for the prediction of current sample. It's originally designed for out-of-distribution detection\cite{devries2018learning}. The trained confidence estimator has two heads, one of which outputs the prediction and the other outputs its confidence. Outliers and noisy labels usually have small confidence value. In our implementation, ResNet-18 is used as backbone and mean teacher mechanism is also adopted. When executing PT algorithm, if some large-loss samples in current mini-batch own large confidence for its prediction, these samples will be reserved and fed to student model. The reserved samples are intuitively 
    clean but hard ones. The confidence threshold is set to 0.9. The accuracy of recognition and confusion matrices are shown in Fig.\ref{confusion_RAF_2}. We can see that the recognition of expressions with much less samples (i.e., fearful) are better with the help of confidence estimator. We list some filtered out images on original RAF-DB in Fig.\ref{filtered_confidence}. It shows good discernment on noisy labels and extreme uncertain samples.

\section{Conclusions}
    In this paper, we propose a semi-supervised framework Progressive Teacher to tackle the shortage of data and inaccurate annotations in facial expression recognition so as to improve the recognition performance. The framework consists of two pairs of teacher-student models. In each pair, the student model computes supervised classification loss and unsupervised consistency loss and then update its parameters with SGD, the teacher model is the average of student model's weight during training process and guides its learning. Auxiliary large-scale unlabeled data is utilized to compute the unsupervised loss. Different from traditional semi-supervised learning method like Mean Teacher, our teachers can select potential clean samples for student models to learn and thus prevent from overfitting noisy samples. Additionally, we use the cross-guidance mechanism to boost performance. Extensive experiments show that our method achieves state-of-the-art result. We obtain the best recognition accuracy of 89.57\% on RAF-DB.

\ifCLASSOPTIONcaptionsoff
  \newpage
\fi



%


\bibliographystyle{IEEEtran}
\bibliography{main}
%
\begin{IEEEbiography}[{\includegraphics[width=1in,height=1.25in,clip,keepaspectratio]{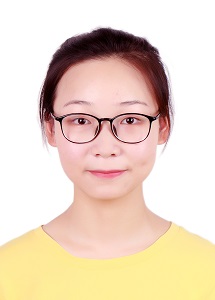}}]{Jing Jiang}
    received the B.E. degree in telecommunication engineering from the Beijing University of Posts and Telecommunications, Beijing, China, in 2020. She is currently pursuing the Ph.D. degree in information and telecommunications engineering. Her research interests include deep learning and facial expression analysis.
\end{IEEEbiography}
\begin{IEEEbiography}[{\includegraphics[width=1in,height=1.25in,clip,keepaspectratio]{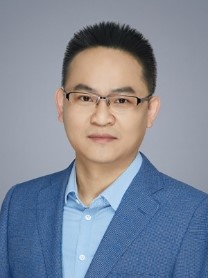}}]{Weihong Deng}
    is a professor in School of Artificial Intelligence, Beijing University of Posts and Telecommunications. His research interests include computer vision and affective computing, with a particular emphasis in face recognition and expression analysis. He has published over 150 technical papers in international journals and conferences, such as IEEE TPAMI, TIP, IJCV, CVPR and ICCV. He serves as area chair for major international conferences such as IJCAI, ACMMM, IJCB, FG, and ICME, and guest editor for IEEE TBIOM, and Image and Vision Computing Journal and the reviewer for dozens of international journals, such as IEEE TPAMI, TIP, TIFS, TNNLS, TAFFC, TMM, IJCV, and PR. His dissertation titled “Highly accurate face recognition algorithms” was awarded the Outstanding Doctoral Dissertation Award by Beijing Municipal Commission of Education in 2011. He has been supported by the program for New Century Excellent Talents in 2014, Beijing Nova in 2016, Young Chang Jiang Scholar, and Elsevier Highly Cited Chinese Researcher in 2020.
\end{IEEEbiography}






\end{document}